\documentclass[journal,twoside,web]{ieeecolor}
\usepackage{generic} 

\usepackage{cite}
\usepackage{amsmath, amssymb, amsfonts}
\usepackage{threeparttable}
\usepackage{booktabs}
\usepackage{algorithmic}
\usepackage{graphicx}
\usepackage{algorithm,algorithmic}
\usepackage{multirow}

\usepackage{hyperref}
\usepackage[caption=false,font=footnotesize]{subfig}

\usepackage{cleveref}
\usepackage[acronym]{glossaries}
\makeglossaries
\newacronym{imu}{IMU}{inertial measurement unit}
\newacronym{ins}{INS}{inertial navigation system}
\newacronym{gnss}{GNSS}{global navigation satellite systems}
\newacronym{slam}{SLAM}{simultaneous localization and mapping}
\newacronym{ekf}{EKF}{extended Kalman filter}
\newacronym{gp}{GP}{Gaussian process}
\newacronym{mains}{MAINS}{magnetic field aided inertial navigation system}
\newacronym{mislam}{MI-SLAM}{magnetic inertial SLAM}
\newacronym{lidar}{LiDAR}{Light Detection and Ranging}
\glsdisablehyper

\newcommand{\ins}{\text{\tiny{ins}}}
\newcommand{\tslam}{\text{\tiny{SLAM}}}

\newcommand{\mains}{\text{\tiny{mains}}}

\def\BibTeX{{\rm B\kern-.05em{\sc i\kern-.025em b}\kern-.08em
    T\kern-.1667em\lower.7ex\hbox{E}\kern-.125emX}}

\begin{document}

\title{MI-SLAM: Magnetic Inertial SLAM Systems}
\author{Chuan~Huang,
        Gustaf~Hendeby,
        and~Isaac~Skog
\thanks{This work has been funded by the Swedish Research Council (Vetenskapsrådet) project 2020-04253 "Tensor-field based localization".}
\thanks{Chuan Huang is with the Dept. of Electrical Engineering and Computer
Science, KTH Royal Institute of Technology, Stockholm, Sweden (e-mail: chuanh@kth.se).}
\thanks{Gustaf Hendeby is with the Dept. of Electrical Engineering, Linköping University (e-mail: gustaf.hendeby@liu.se).}
\thanks{Isaac Skog is with 
the Dept.of Electrical Engineering and Computer
Science, KTH Royal Institute of Technology, 10044 Stockholm, Sweden, and
also with the Division of Underwater Technology, Swedish Defence Research
Agency (FOI), Kista, Sweden (e-mail: skog@kth.se).}}
\maketitle

\begin{abstract}
Spatially inhomogeneous magnetic fields offer a valuable, non-visual information source for positioning. Among systems leveraging this, magnetic field-based \gls{slam} systems are particularly attractive. These systems execute positioning and magnetic field mapping tasks simultaneously, and they have bounded positioning error within previously visited regions. However, state-of-the-art magnetic-field \gls{slam} methods typically require low-drift odometry data provided by visual odometry, a wheel encoder, or pedestrian dead-reckoning technology, which may not be available under certain situations. To address this limitation, this work proposes two \gls{mislam} systems—a loosely coupled version and a tightly coupled version—that rely solely on low-cost sensors: an \gls{imu} and seven magnetometers. Both systems use a \gls{mains} to obtain odometry. The key difference between the two systems is whether the navigation state estimation is done in one or two steps. These systems are evaluated in real-world indoor environments with multi-floor structures. The proposed \gls{mislam} systems outperform the state-of-the-art magnetic-inertial odometry system, \gls{mains}, achieving a positioning RMSE below 1 meter on a trajectory lasting approximately 200 seconds. A potential application of the proposed systems is for the positioning of emergency response officers in mine or fire rescue operations, where one cannot rely on a GNSS or visual-based localization system.
\end{abstract}




\section{Introduction}
\IEEEPARstart{P}{ositioning} and navigation services are essential for various applications, including scientific research, emergency response, and military operations. These services have historically relied on positioning information from \gls{gnss}.
However, GNSS-denied environments, such as indoors or subsurface, where GNSS signals are unavailable or unreliable, pose significant challenges for accurate and reliable positioning.
Therefore, developing robust and accurate positioning systems that can operate effectively in GNSS-denied environments is of paramount importance.

A common approach to address the challenges of positioning in GNSS-denied environments is to use additional sensors, such as cameras, \gls{lidar}, and inertial measurement units (\gls{imu}s), to complement the \gls{gnss}. 
For example, visual and \gls{lidar} \gls{slam} systems~\cite{Mur-Artal2015ORB-SLAM,Hess2016Real-time,Zhang2025RealTime,Zhang2023VisualInertial} have been widely used for indoor positioning. However, these systems have their own limitations. Visual SLAM systems are sensitive to lighting conditions and may fail in low-light or featureless environments~\cite{Park2017Illumination}, and LiDAR SLAM systems can be expensive and may function poorly in environments with limited geometric features~\cite{Zou2022Comparative}. 
Thus, it is necessary to exploit other types of environmental features to achieve robust indoor \gls{slam}.

Spatial variations of indoor magnetic fields have been recognized as a promising source of environmental features for \gls{slam}~\cite{Li2012Magnetic}. Because of the ubiquitous presence of ferromagnetic materials in buildings, the indoor magnetic field exhibits spatial variations over short distances. Moreover, indoor magnetic fields can be measured using low-cost magnetometers, which are commonly found in smartphones and other portable devices. This makes magnetic field \gls{slam} systems cost-effective and accessible.

Previous works have shown that magnetic field \gls{slam} systems can achieve meter-level positioning accuracy in indoor environments~\cite{Kok2018MagSlam,Viset2022EKF,Vallivaara2011Magnetic,pavlasek2023magnetic,Vallivaara2026Saying}. However, most of these systems rely on low-drift odometry to maintain accurate localization while exploring previously unknown environments before loop closure occurs. Such odometry is typically obtained from wheel encoders, pedestrian dead reckoning, or visual-inertial odometry. These approaches, however, have inherent limitations: wheel encoders are restricted to wheeled platforms operating on planar surfaces, pedestrian dead reckoning is specific to human locomotion, and visual-inertial odometry inherits many of the limitations of visual \gls{slam}, such as sensitivity to lighting conditions and texture-poor environments. Consequently, these constraints limit the applicability of magnetic field \gls{slam} systems.

One possible solution is to develop magnetic field \gls{slam} systems that use only \gls{imu} and magnetometers, because these two types of sensors combined are sufficient, in theory, to achieve \gls{mislam}. The \gls{imu} can be used for odometry and does not suffer from those constraints. However, the accuracy of odometry derived from an \gls{imu} depends on its grade. For example, the accumulated error in integrating odometric data derived from low-cost consumer-grade \gls{imu}s is typically on the order of hundreds of meters after one minute of navigation and grows at a rate of elapsed time cubed; thus, low-cost consumer-grade \gls{imu}s alone cannot provide low-drift odometry for any practically useful magnetic field \gls{slam} system. It must work with magnetometers. In a recent work~\cite{huang2023mains}, a magnetic-field-aided inertial navigation system (MAINS) that uses a low-cost \gls{imu} and an array of 30 magnetometers was developed. \gls{mains} uses the magnetometer array to measure variations in local magnetic fields and uses this information and IMU data to infer pose change when the navigation platform moves. This system is essentially a magnetic-inertial odometry system that could potentially replace the odometry used in existing magnetic field SLAM systems.  

Inspired by \gls{mains}, we seek to answer the following research questions:
\begin{itemize}
    \item Can a 3D magnetic inertial SLAM (\gls{mislam}) system be built using only a low-cost IMU and a magnetometer array?
    \item How does the performance of a tightly coupled \gls{mislam} system compare to a loosely coupled \gls{mislam} system?
    \item How do height measurements from a barometer and IMU measurements of different qualities affect the overall performance of the SLAM system?
\end{itemize}
To that end, we propose two \gls{mislam} systems: a loosely coupled and a tightly coupled system. We evaluate the proposed systems on real-world datasets to answer these research questions.
\subsection{Related Work}
Many methods for magnetic-field-based indoor positioning have been proposed. They are generally classified into two categories: infrastructure-based methods~\cite{Prigge2004Signal} and infrastructure-free methods~\cite{Wang2024Orientation,Shen2024IDFMFL}. The former generally requires some magnetic source to be installed in the environment, such as magnetic coils or magnets. The latter, on the other hand, exploits the ambient magnetic field as a location fingerprint. Infrastructure-free methods are often preferred because they eliminate the need for dedicated infrastructure, thereby reducing deployment and maintenance efforts.

The infrastructure-free methods can be further classified into two categories: fingerprint-based methods\cite{ZHANG2023MagODO,Wang2024CrowdMagMap,Kuang2025CrowdMagMap,Kok2024Online,Shen2024IDFMFL,Chen2025APP} and model-based methods~\cite{skog2018magnetic,skog2021magnetic,huang2023mains,Wang2024Orientation,Viset2022EKF,zmitri2020magnetic,Vallivaara2026Saying,Kok2018MagSlam}. The fingerprint-based methods are based upon the assumption that the magnetic fields at different locations are distinct. These methods generally rely on matching magnetic field measurements to a pre-established magnetic field map to estimate position, or to a stored measurement history to recognize previously visited locations for loop closure. The model-based methods, on the other hand, seek to use a mathematical model to describe magnetic fields measurements. These methods can use the magnetic measurements to update their magnetic field model (build a magnetic field map) and estimate pose change as the device moves through the environment. There are pros and cons for both methods. The fingerprint-based methods can achieve high accuracy if magnetic field matching is successful, but they require carefully designed feature extraction and matching algorithms. Furthermore, erroneous matching can lead to significant positioning errors. In contrast, the model-based methods are easier to establish a theoretically sound framework for the inference process, but they can be sensitive to the choice of the magnetic field model and the tuning of the model parameters.

The proposed \gls{mislam} systems are model-based. They were made possible by developments in magnetic-inertial odometry and previous work on magnetic field modeling. In~\cite{skog2018magnetic}, a polynomial model was proposed to locally approximate the magnetic field, which was later extended in~\cite{skog2021magnetic} to estimate pose changes of a magnetometer array. This polynomial model was also used in~\cite{huang2023mains} to build a \gls{mains}, which effectively reduced the positioning drift of the inertial navigation system by two orders of magnitude. This work~\cite{huang2023mains} constitutes one of the building blocks of the proposed systems. The other building block comes from the work in global magnetic field modeling. Originally proposed in~\cite{Solin2018Modeling}, the reduced rank~\gls{gp} model was shown to be effective and computationally efficient in modeling the indoor magnetic field on a global scale. The GP model was later used in \cite{viset2021Magnetic,Kok2018MagSlam,Viset2022EKF} to build magnetic field SLAM systems. 

\subsection{Contributions}
The contributions of this work are as follows:
\begin{itemize}
     \item We propose a loosely coupled and a tightly coupled 3D \gls{mislam} system. These two system architectures serve as practical reference designs for practitioners.
    \item We conduct an evaluation of the proposed systems in terms of their positioning performance on real-world datasets, including an indoor environment with multi-floor structures. 
\end{itemize}

\textbf{Reproducible research}: Both the datasets and code used in this study are available at \href{https://github.com/Huang-Chuan/MI-SLAM}{https://github.com/Huang-Chuan/MI-SLAM}.

\section{Preliminaries}
As magnetic field modeling is essential for magnetic field SLAM, we briefly review two commonly used magnetic field models: the first captures local, small-scale variations, while the second describes global, large-scale fields.

In the absence of time-varying electric fields and current density in the region of interest, the magnetic field fulfills the curl-free condition~\cite[p.~180]{jackson2021classical}
\begin{equation}
    \nabla \times M(r) = 0
\end{equation}
where $r\in\mathbb{R}^3$ denotes the position in the Cartesian coordinates. This permits the expression of $M(r)$ as the gradient of a scalar potential function $\phi(r):\mathbb{R}^3\rightarrow \mathbb{R}$, i.e.,
\begin{equation}
\label{eq: scalar potential}
    M(r) = \nabla \phi(r).
\end{equation}
Both magnetic field models presented next are based on~\eqref{eq: scalar potential}.

\subsection{Local Magnetic Field Model}  \label{subsec: polynomial model}
A local magnetic field model refers to a model that captures the magnetic field in a small region of space where the field can be well approximated by a simple function. The polynomial model proposed in~\cite{skog2018magnetic} is a model of this type. It provides a compact local representation of the magnetic field. In this model, the scalar potential function is assumed to be an $(n+1)$-degree polynomial function, i.e., 
\begin{equation}
\phi(r;c) =
\;\sum_{\substack{i,j,k \ge 0 \\ i+j+k \le n+1}}
   c_{i,j,k} \, r_x^i r_y^j r_z^{k}.
\end{equation}
Here $c_{i,j,k} \in \mathbb{R}$ denotes the coefficient and $r\triangleq[r_x\;r_y\;r_z]^\top$ denotes the position.
Furthermore, the coefficients of the resulting magnetic field model $M(r;c)$ from applying \eqref{eq: scalar potential} are further constrained by the divergence-free condition~\cite[p.~180]{jackson2021classical}
\begin{equation}
    \nabla \cdot M(r;c) = 0.
\end{equation}
It leads to a $n$-th degree polynomial magnetic field model~\cite{skog2018magnetic}
\begin{equation}
\label{eq: polynomial model}
    M(r;\theta) = \Phi(r)\theta
\end{equation}
where $\Phi(r)\in\mathbb{R}^{3\times N_p}$ denotes the regressor matrix, $\theta \in \mathbb{R}^{N_p \times 1}$ denotes the coefficients, with $N_p = n^2+4n+3$.

It has been empirically verified in \cite{huang2023mains} that a first-degree polynomial magnetic field model can capture indoor magnetic fields on a planar magnetometer array of size $345\, \text{mm} \times 245 \, \text{mm}$. The regressor matrix in this case is 
\begin{subequations}
    \label{eq: regressor and theta}
\begin{equation} \label{eq: regressor matrix}
    \Phi(r)=\begin{bmatrix} 
         1 & 0& 0&  0&     0& r_z & r_y &  2r_x\\
            0& 1& 0& r_z&  2r_y&  0& r_x&     0\\
            0& 0& 1& r_y& -2r_z& r_x&  0& -2r_z
    \end{bmatrix}
\end{equation}
and 
\begin{equation}
    \theta=\left[\theta(1) \; \theta(2)\; \cdots\;\theta(8)\right]^{\top}.
\end{equation}
\end{subequations}

\subsection{Global Magnetic Field Model} \label{subsec: gp model}
A global magnetic field model describes the magnetic field throughout an environment, such as a building or a large lab. By modeling the scalar magnetic potential $\phi(r)$ as a \gls{gp} and approximating it using Hilbert space methods, Solin et al.~\cite{Solin2018Modeling} proposed a global magnetic field model for magnetic fields in a cuboid of dimension ($[-U_x,U_x]\times[-U_y,U_y]\times[-U_z,U_z]$), and the model is given by
\begin{subequations}    \label{chap: background eq: global magnetic field model}
\begin{equation} 
M(r;\eta) = \nabla \Psi(r)\eta
\end{equation}
where
\begin{align}
    \Psi(r)&=[r^{\top}\;\psi_1(r)\;\psi_2(r)\;\cdots\;\psi_{N_m}(r)]\\ 
    \psi_j(r)&=\prod_{d=x,y,z}\frac{1}{\sqrt{U_d}}\sin\left(\frac{\pi n_{j,d}(r_d+U_d)}{2U_d}\right)\\
    \eta &\sim \mathcal{N}(0, \Lambda)\\
\Lambda=\text{diag}&\left(\sigma_{\text{lin}}^2,\sigma_{\text{lin}}^2,\sigma_{\text{lin}}^2,S_{\text{SE}}(\lambda_1),S_{\text{SE}}(\lambda_2),\cdots,S_{\text{SE}}(\lambda_{N_m})\right)\\
    \lambda_j^2&=\sum_{d=x,y,z}\left(\frac{\pi n_{j,d}}{2U_d}\right)^2\\
S_{\mathrm{SE}}(\lambda_j)&=\sigma_{\mathrm{SE}}^2\left(2 \pi \ell_{\mathrm{SE}}^2\right)^{3 /2} \exp \left(-\frac{\lambda_j^2 \ell_{\mathrm{SE}}^2}{2}\right).
\end{align}    
\end{subequations}
Here $\nabla$ denotes the gradient operator and $\{\psi_j(r)\}$ are the basis fucntions. $\mathcal{N}(0, \Lambda)$ denotes a random variable with zero mean and covariance matrix $\Lambda$. Furthermore, $\sigma_{\text{lin}}$ denotes the magnitude scale of the linear kernel, and $\sigma_{\text{SE}}$ and $l_{\text{SE}}$ denote the magnitude scale and the characteristic length scale of the square exponential kernel, respectively. Lastly, $n_{j,d}\in\mathbb{Z}^{+}$ denotes the element of an $N_m\times3$ matrix indexed at $(j,d)$, and the details of constructing the matrix are presented in~\cite{Solin2018Modeling}.

\subsection{Discussion}
The polynomial model presented in Section~\ref{subsec: polynomial model} is well-suited for local modeling, due to its simplicity and computational efficiency. The first-degree polynomial model was used in~\cite{huang2023mains} to model the magnetic field within the small area covered by the magnetometer array at each time step. In this model, both the spatial coordinates and the magnetic field are typically expressed in the body frame of the magnetometer array. 

On the other hand, the approximated \gls{gp} model presented in Section~\ref{subsec: gp model} is appropriate for global modeling, as the sinusoidal basis functions can capture the magnetic field variations across a larger area. Typically, several hundreds to thousands of basis functions are used to represent the magnetic field across an area of several hundred square meters~\cite{Viset2022EKF}. Therefore, this model is computationally costly. Since this model aims to describe the field over a global region, both the spatial coordinates and the magnetic field are typically expressed in the navigation frame. 

\section{Magnetic-Field-Based Positioning Systems}
In this section, we briefly review the system models used in two magnetic field-based positioning systems. The first model is used by the magnetic field SLAM system in~\cite{Viset2022EKF}, and the second by \gls{mains} in~\cite{huang2023mains}.

\subsection{The Magnetic Field SLAM System}\label{subsec: EKF-SLAM}
The state-of-the-art magnetic field SLAM system~\cite{Viset2022EKF} uses low-drift odometry and magnetic field measurements from a single magnetometer. The state vector of the magnetic field SLAM at time step $k$ is defined as $x_{k}^{\tslam} \triangleq [p_{k}^{\top}\;q_{k}^{\top}\; \eta_{k}^{\top}]^\top$, where $p_{k}\in\mathbb{R}^3$ denotes the position, $q_{k}\in S^3$ denotes the orientation in quaternion representation, and $\eta_{k}\in\mathbb{R}^{3+N_m}$ denotes the coefficients of the approximated \gls{gp} magnetic field model.

The system model of the magnetic field SLAM is
\begin{subequations}
\label{E: SLAM SSM}
    \begin{align}
    \label{E: SLAM SSMa}
    x_{k+1}^{\tslam} & = f^{\tslam}(x_{k}^{\tslam}, \tilde{u}_{k}^{o})\\
    y_{k}^{(0)}&= h^{\tslam}(x_{k}^{\tslam}) +e_{k}^{(0)}
\end{align}
where
\begin{equation}
f^{\tslam}(x_{k}^{\tslam}, \tilde{u}_{k}^{o}) =\begin{bmatrix}
p_{k} + \Delta p_{k}  \\ 
q_{k} \otimes \Delta q_{k} \\
    \eta_{k} 
\end{bmatrix}
\end{equation}
and
\begin{equation} \label{eq: original magSLAM measurement equation}
h^{\tslam}(x_{k}^{\tslam})
= R_{k}^{\top} \nabla\Psi\big(p_{k}\big) \eta_{k}.
\end{equation}
\end{subequations}
Here $\tilde{u}_{k}^{o}=[\Delta p_{k}^{\top} \;\Delta q_{k}^{\top}]^\top$ denotes the visual odometry data, where $\Delta p_{k}\in\mathbb{R}^3$ and $\Delta q_{k}\in S^3$ denote the change in position and orientation, respectively. Furthermore, $y_k^{(0)}\in \mathbb{R}^3$ and $e_k^{(0)}\in \mathbb{R}^3$ are the magnetometer measurements and the Gaussian white measurement noise, respectively. In addition, $\otimes$ denotes quaternion multiplication, and $R_{k}\in SO(3)$ denotes the rotation matrix corresponding to the quaternion $q_{k}$. 

\subsection{The Magnetic-Field-Aided Inertial Navigation System}\label{subsec: MAINS}
\gls{mains} proposed in~\cite{huang2023mains} uses a low-cost \gls{imu} and a magnetometer array to provide low-drift inertial navigation. The state vector of MAINS at time step $k$ is defined as
\begin{subequations}
\begin{equation}
    x_k^{\mains} \triangleq [{x_k^{\ins}}^\top  \; \theta_k^\top]^\top
\end{equation}
\begin{equation}
    x_k^{\ins} \triangleq [p_k^{\top}\; v_k^{\top}\; q_k^{\top}\; b_{a,k}^{\top}\; b_{g,k}^{\top}]^\top
\end{equation}
\end{subequations}
where $x_k^{\ins}$ denotes the inertial navigation states. Furthermore, $v_k\in\mathbb{R}^3$ denotes the velocity, $b_{a,k}\in\mathbb{R}^3$ and $b_{g,k}\in\mathbb{R}^3$ denote the accelerometer and gyroscope biases, respectively, and $\theta_k\in\mathbb{R}^{N_p}$ denotes the coefficients of the polynomial magnetic field model presented in~\Cref{subsec: polynomial model}. The measurement $y_k\in\mathbb{R}^{3N}$ is a stacked vector containing the measurements from all $N$ magnetometers in the array, i.e., 
\begin{equation}
    y_k \triangleq \left[ \big(y_k^{(1)}\big)^{\!\top} \; \big(y_k^{(2)}\big)^{\!\top} \; \cdots \; \big(y_k^{(N)}\big)^{\!\top} \right]^{\!\top}
\end{equation}
where $y_k^{(i)}\in\mathbb{R}^3$ denotes the measurements from the $i\text{-th}$ magnetometer.
The dynamics of the state vector and the measurement model are
\begin{subequations}
\label{E: SSM}
	\begin{align}
	\label{E: SSMa}
	x_{k+1}^{\mains} & = f^{\mains}(x_k^{\mains},\tilde{u}_k, w_k)\\
    \label{E: mains measurement equation}
	y_k &= H^{\mains} x_k^{\mains} +e_k
\end{align}
where
\begin{equation}
f^{\mains}(x_k^{\mains},\tilde{u}_k, w_k) =\begin{bmatrix}
 f^{\ins}(x^{\ins}_k,\tilde{u}_k, w^{\ins}_k) \\
f^{\theta}(\theta_k, x^{\ins}_k,\tilde{u}_k, w^{\theta}_k)
\end{bmatrix}
\end{equation}
and
\begin{equation} \label{eq: mains meas equation}
H^{\mains}= \begin{bmatrix}
        0_{3\times 16} & \Phi(r^{(1)})\\
        \vdots &\vdots\\
        0_{3\times 16} &\Phi(r^{(N)})\\
    \end{bmatrix}.
\end{equation}
\end{subequations}
Here $f^{\ins}(\cdot)$ denotes the inertial subsystem dynamics, $f^{\theta}(\cdot)$ denotes the magnetic subsystem function. The exact form of these functions is presented in~\cite{huang2023mains}. Moreover, $\tilde{u}_k=[\tilde{a}_k^{\top}\;\tilde{\omega}_k^{\top}]^\top$ denotes the IMU measurements, where $\tilde{a}_k\in\mathbb{R}^3$ and $\tilde{\omega}_k\in\mathbb{R}^3$ denote the specific force and angular velocity measurements, respectively. Furthermore, $w_k=[(w_k^{\ins})^{\top}\;(w_k^{\theta})^{\top}]^\top$ denotes the process noise, where $w_k^{\ins}\in\mathbb{R}^{12}$ and $w_k^{\theta}\in\mathbb{R}^{N_p}$ denote the process noise of the inertial and magnetic subsystems, respectively. $r^{(i)}\in\mathbb{R}^3$ denotes the position of the $i$-th magnetometer in the array expressed in the body frame. 
$\Phi(\cdot)$ denotes the regressor matrix in~\eqref{eq: polynomial model}. Finally, $e_k \in \mathbb{R}^{3N}$ denotes the stacked white Gaussian measurement noise.

\begin{figure}
    \centering
    \subfloat[]{%
        \includegraphics[width=0.6\columnwidth]{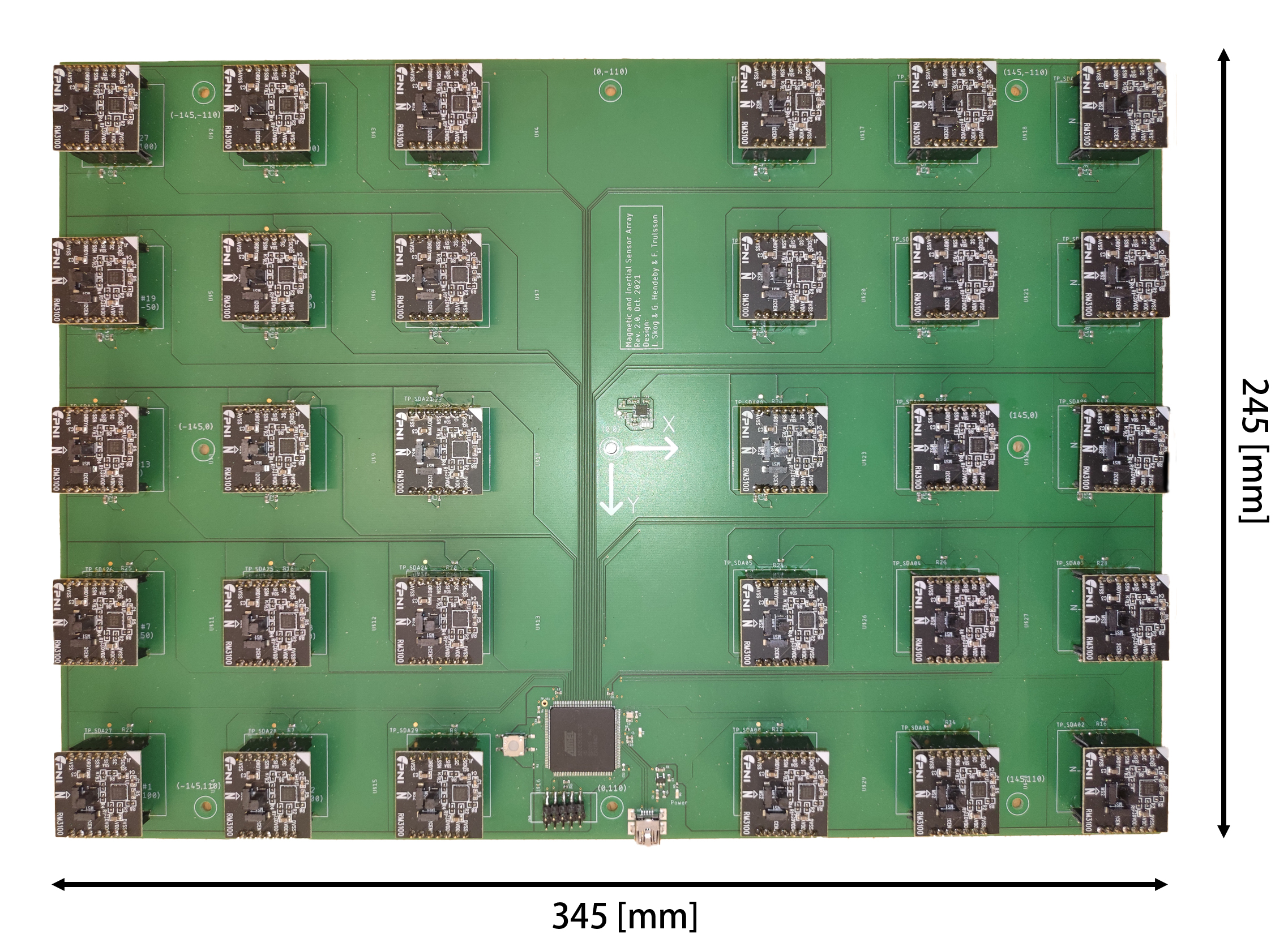}%
    }
\hfill
    \subfloat[]{%
        \includegraphics[width=0.35\columnwidth]{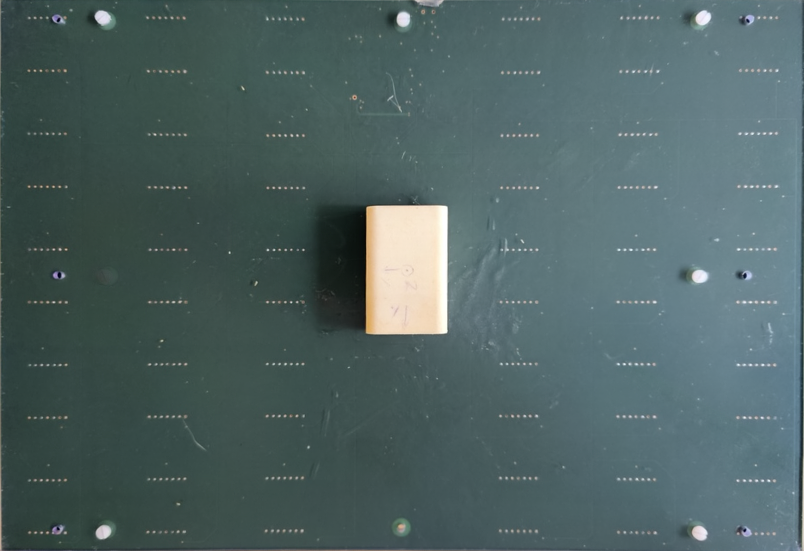}%
    }
\caption{The sensor board used in \gls{mains}. It has 30 PNI \href{https://www.pnicorp.com/rm3100/}{RM3100} magnetometers and an \href{https://www.inertialelements.com/osmium-mimu4444.html}{Osmium MIMU 4844} IMU mounted on the bottom side. (a) Front side of the sensor board showing the magnetometer array. (b) Back side of the sensor board showing the \gls{imu}.}
\label{fig: sensor board}
\end{figure}

\subsection{Summary}
Both the magnetic field SLAM system and the \gls{mains} have their own advantages and limitations. The magnetic field SLAM system has a bounded positioning error when in previously mapped areas. However, its reliance on the low-drift odometry limits its usability. \gls{mains}, on the other hand, can provide low-drift inertial navigation, but it only maintains a local magnetic field map; thus, positioning errors grow without bound.
In the next section, we propose a loosely coupled magnetic field SLAM system that combines the advantages of both systems while mitigating their limitations.

\section{The Loosely Coupled MI-SLAM System}
The loosely coupled MI-SLAM system integrates MAINS~\cite{huang2023mains} with a magnetic field SLAM subsystem in a cascaded configuration. It operates across two distinct time scales, following a common paradigm in visual SLAM systems, i.e., a high-rate odometry and a low-rate mapping process~\cite{carlone2025slam}. 

In our implementation, MAINS runs at the \gls{imu} rate and is indexed by $k$, whereas the modified magnetic field SLAM system operates at a rate downsampled by a factor of $K$ and is indexed by $n$. The relationship between the two time indices is given by $k = 1 + nK$. 

This design reduces the overall computational load, as the magnetic field SLAM subsystem is more computationally intensive than the \gls{mains} subsystem, typically having a state dimension an order of magnitude larger than that of MAINS.

\subsection{Computing Odometry Data with MAINS}\label{subsec: odometry}
MAINS uses the error-state Kalman filter~\cite{Joan2017Quaternion} to estimate the current pose. However, odometry computation involves both the current and past pose estimates. Mathematically, the odometric data between
the time step $i = 1 + nK$ and $j = 1 + (n + 1)K$ can be computed as
\begin{subequations}
\begin{equation}
    \tilde{u}^o_{n} = \begin{bmatrix}
        \Delta \hat{p}_{i,j} \\
        \Delta \hat{q}_{i,j}
    \end{bmatrix} = 
    \begin{bmatrix}
        \hat{p}_{j}- \hat{p}_{i}\\
        \hat{q}_{i}^{\ast}\otimes \hat{q}_{j}
    \end{bmatrix}     
\end{equation}
\begin{equation} \label{eq: odometry cov}
    \text{var}(\tilde{u}_{n}^o) = 
        \begin{bmatrix}
            \text{var}(\Delta \hat{p}_{i,j}) & \text{K}_{\Delta \hat{p}_{i,j}\Delta \hat{q}_{i,j}} \\
            \text{K}_{\Delta \hat{q}_{i,j} \Delta \hat{p}_{i,j}} & \text{var}(\Delta \hat{q}_{i,j})
        \end{bmatrix}.
\end{equation}
\end{subequations}
Here $\Delta \hat{p}_{i,j}$ and $\Delta \hat{q}_{i,j}$ denote the translational displacement and orientation change estimates during the time interval $(i,j)$. $\hat{p}_{i}$ and $\hat{q}_{i}$ denote the position and orientation estimates at time $i$, respectively. Furthermore, $q^{\ast}$ denotes the quaternion conjugation. Moreover, $\text{var}(\cdot)$ and $\text{K}_{\cdot\cdot}$ denote the covariance and cross-covariance matrix of the random vector(s), respectively. The computation of these matrices are detailed in the appendix.

\subsection{The Magnetic Field SLAM Subsystem}
The magnetic field SLAM subsystem is largely the same as the one presented in Section~\ref{subsec: EKF-SLAM}, except that the measurement equation~\eqref{eq: original magSLAM measurement equation} is modified to accommodate magnetometer(s) not placed at the center of the body frame. Precisely, the measurement equation for the $i$-th magnetometer in the array is
\begin{subequations} \label{eq: magslam equation}
    \begin{equation}
        y_{n}^{(i)}= h^l(x_{n}^{\tslam};r^{(i)}) +e_{n}^{(i)}
    \end{equation}
where
\begin{equation}h^{l}(x_{n}^{\tslam})
= 
R_{n}^{\top} \nabla\Psi\big(p_{n} + R_{n} r^{(i)}\big)  \eta_{n}.
\end{equation}
\end{subequations}
Here $e_n^{(i)}\in\mathbb{R}^3$ denotes the white Gaussian measurement noise.

\subsection{Summary}
The loosely coupled MI-SLAM system is a straightforward implementation built on the magnetic field SLAM system and \gls{mains}. A key advantage of this architecture is that its computational complexity can be readily controlled by adjusting the odometry computation frequency, which in turn determines the measurement update frequency of the magnetic field SLAM subsystem. Since the measurement update in the magnetic field SLAM subsystem is the most computationally expensive operation, reducing its update frequency effectively lowers the overall computational cost.
However, this architecture has an inherent limitation: the global and local magnetic field models are maintained in separate systems, and their correlation is therefore neglected. As a result, the magnetic field map is constructed using only a subset of the available magnetic field measurements due to the reduced update frequency. To overcome this limitation, we propose a tightly coupled MI-SLAM system in the next section.

\section{The Tightly Coupled MI-SLAM System} \label{sec: tightly-coupled SLAM}
The tightly coupled MI-SLAM system integrates both the \gls{mains} and magnetic field \gls{slam} systems into a unified SLAM framework. In this framework, the state vector encompasses the full set of states, including the inertial navigation states and the local and global magnetic field model coefficients. Next, the state-space model of this system is presented.

\subsection{State Dynamics}
The state vector of the tightly coupled SLAM system at time step $k$ is defined as $ x_k \triangleq [p_k^{\top}\;v_k^{\top}\;q_k^{\top}\;b_{a,k}^{\top}\;b_{g,k}^{\top}\; \theta_k^{\top}\; \eta_k^\top]^\top$, and the state transition model is
\begin{subequations}
\begin{equation}
\label{E: Tightly SSM}
    x_{k+1}  = f(x_k,\tilde{u}_k, w_k)
\end{equation}
where   
\begin{equation}
f(x_k,\tilde{u}_k, w_k) =\begin{bmatrix}
 f^{\ins}(x^{\ins}_k,\tilde{u}_k, w^{\ins}_k) \\
f^{\theta}(\theta_k, x^{\ins}_k,\tilde{u}_k, w^{\theta}_k) \\
\eta_k
\end{bmatrix}.
\end{equation}
\end{subequations}
\subsection{Measurement Equations}
Since the state vector includes both the local and global magnetic field model coefficients, the magnetometer array's measurements can be expressed in terms of both models. To exploit their complementary advantages, one can use the measurement equations in \gls{mains} or the magnetic field SLAM system in an alternating manner. A natural choice is to use the computationally efficient measurement equation from \gls{mains} most of the time, while occasionally using the more complex measurement equation from the magnetic field SLAM system.
In this way, the system can benefit from the low-drift inertial navigation provided by the \gls{mains} while still leveraging the global consistency of the magnetic field SLAM system.
In the proposed tightly coupled MI-SLAM system, the measurement equation of the global model, which is based on~\eqref{eq: magslam equation}, is further simplified to reduce computational complexity. The basic idea is that the magnetic field at any sensor's location can be viewed as a superposition of the magnetic field in the center of the array and some local magnetic field variation. This results in 
\begin{subequations} \label{eq: mixed}
    
\begin{equation}
    y_k^{(i)} = h^{t}(x_k;r^{(i)}) + e_k^{(i)}
\end{equation}
where
\begin{equation} \label{eq: fused measurement model}
h^{t}(x_k;r^{(i)}) = R_k^{\top}\nabla\Psi\big(p_k\big)\eta_k + \left( \Phi(r^{(i)})-\Phi(0) \right)\theta_k.
\end{equation}
\end{subequations}
Here, the first term computes the magnetic field at the center of the magnetometer array by the approximated \gls{gp} model, while the second term captures the local variations across the array modeled by the polynomial model. 

To sum up, there are two measurement equations for the $i$-th magnetometer measurements: the equation~\eqref{eq: mains meas equation} used by~\gls{mains} and the equation~\eqref{eq: mixed}. They are used in an alternating manner: the former is employed at every update step except once every $D$ samples, when the second model is used instead. That is,
\begin{equation} \label{eq: tightly imslam model}
 y_k^{(i)} =
  \begin{cases}
  h^t(x_k;r^{(i)}) + e_k^{(i)}  & \text{for}\; k = D, 2D, \cdots \\
  \left[0\; \Phi\left(r^{(i)}\right)\; 0\right] x_k + e_k^{(i)}   & \text{otherwise}
  \end{cases}
\end{equation}

\subsection{Summary}
The tightly coupled \gls{mislam} system augments the \gls{ins} state with both local and global magnetic field model coefficients, performing measurement updates with these models in an alternating fashion. By having two magnetic field models within a single state-space model, the correlation between these models can be utilized in magnetic field mapping. Furthermore, by estimating the navigation state in a single step, the system bypasses the computation of odometry data described in \ref{subsec: odometry} in the loosely coupled MI-SLAM. However, the computational demand of the tightly coupled \gls{mislam} system remains significantly higher than that of the loosely coupled \gls{mislam} system. This is primarily because the state dimension of the tightly coupled \gls{mislam} system is comparable to that of the magnetic field \gls{slam} subsystem in the loosely coupled system, while its measurement update is performed at the \gls{imu} rate.

\section{Experiment}
\begin{figure}
    \centering
    \includegraphics[width=0.8\columnwidth]
    {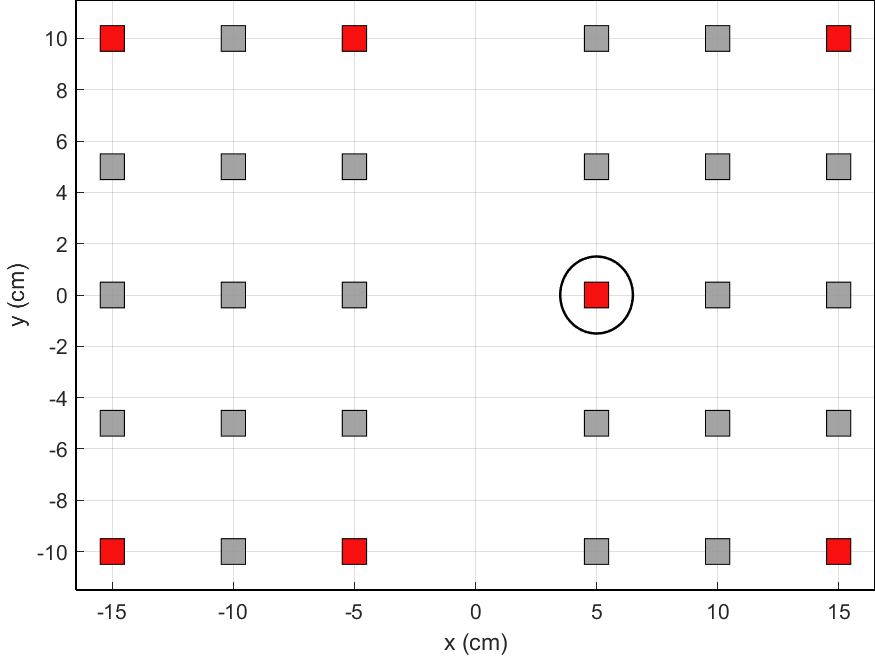}
    \caption{The seven magnetometers (in red) used in the experiments. In the loosely coupled \gls{mislam} systems, all seven magnetometer measurements were used in~\cref{E: mains measurement equation} while the circle-marked magnetometer measurement was used in~\cref{eq: magslam equation}. In the tightly coupled \gls{mislam} systems, all seven magnetometer measurements were used in the second branch of~\cref{eq: tightly imslam model} while the circle-marked magnetometer measurement was used in the other branch.}
    \label{fig: sensor configuration}
\end{figure}

To evaluate the performance of the proposed loosely coupled and tightly coupled \gls{mislam} systems, we conducted experiments in an indoor environment using a custom-built sensor array, see Fig.~\ref{fig: sensor board}. The sensor array is equipped with an IMU and 30 magnetometers. The size of the array is $345\,\text{mm}\times245\,\text{mm}$. In the experiments, however, only seven magnetometers were used (see Fig.~\ref{fig: sensor configuration}) to reduce magnetic interference between closely spaced sensors and to lower the hardware cost. Furthermore, the IMU and magnetometer array are jointly calibrated using the method proposed in~\cite{Huang2026Joint}.

\subsection{Proof-of-Concept Experiment} \label{subsec: proof-of-concept}
In this experiment, we held the sensor array and walked five laps around an $8\,\text{m}\times8\,\text{m}$ square in a room equipped with a motion capture system. The \gls{imu} and magnetometers were sampled at 100 Hz, and the resulting dataset has a duration of 165 seconds.

We selected the state-of-the-art magnetic inertial odometry system, \gls{mains}~\cite{huang2023mains}, in comparison to the two proposed \gls{mislam} systems. All three systems used the tracked poses by a motion capture system at the beginning of the trajectory for $20$ seconds. This is to stabilize the IMU bias estimates, which is important for good positioning performance~\cite[Ch. 22]{braasch2023fundamentals}. During the later parts of the trajectory, these systems relied solely on measurements from the onboard sensors for positioning. In the loosely coupled MI-SLAM system, the odometry data were internally computed at 5 Hz ($K\!=\!20$). In the tightly coupled SLAM system, the magnetic field model switching time was also set to 0.2 seconds ($D\!=\!20$) to match the odometry data computation frequency. The length scale of the approximated \gls{gp} model $l_{\text{SE}}$ was set to $1$ meter, and the number of basis functions $N_m$ was set to 80.

\begin{figure*}
    \centering
    \subfloat[]{%
        \includegraphics[width=0.29\textwidth]{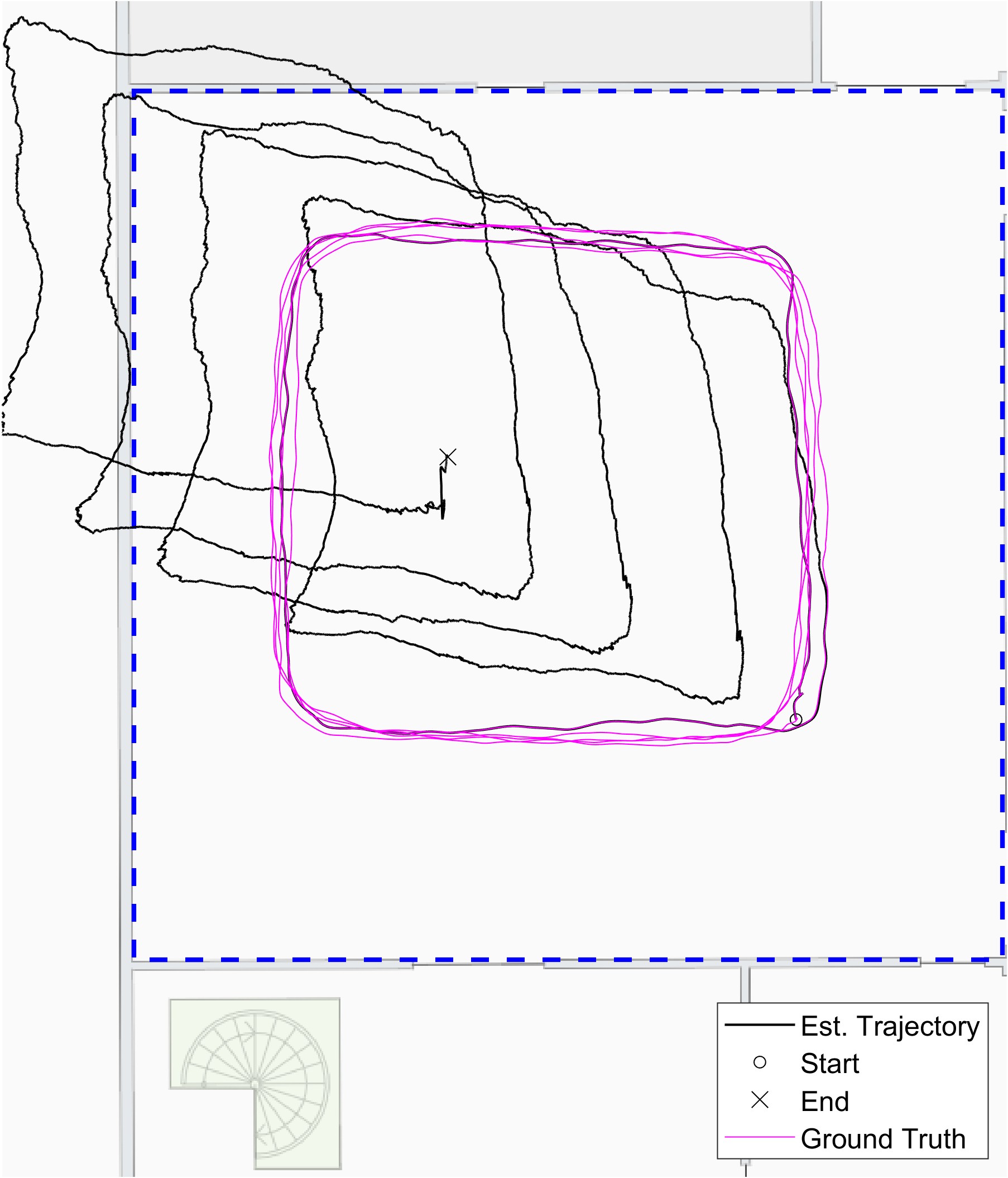}%
    }
\hfill
    \subfloat[]{%
        \includegraphics[width=0.33\textwidth]{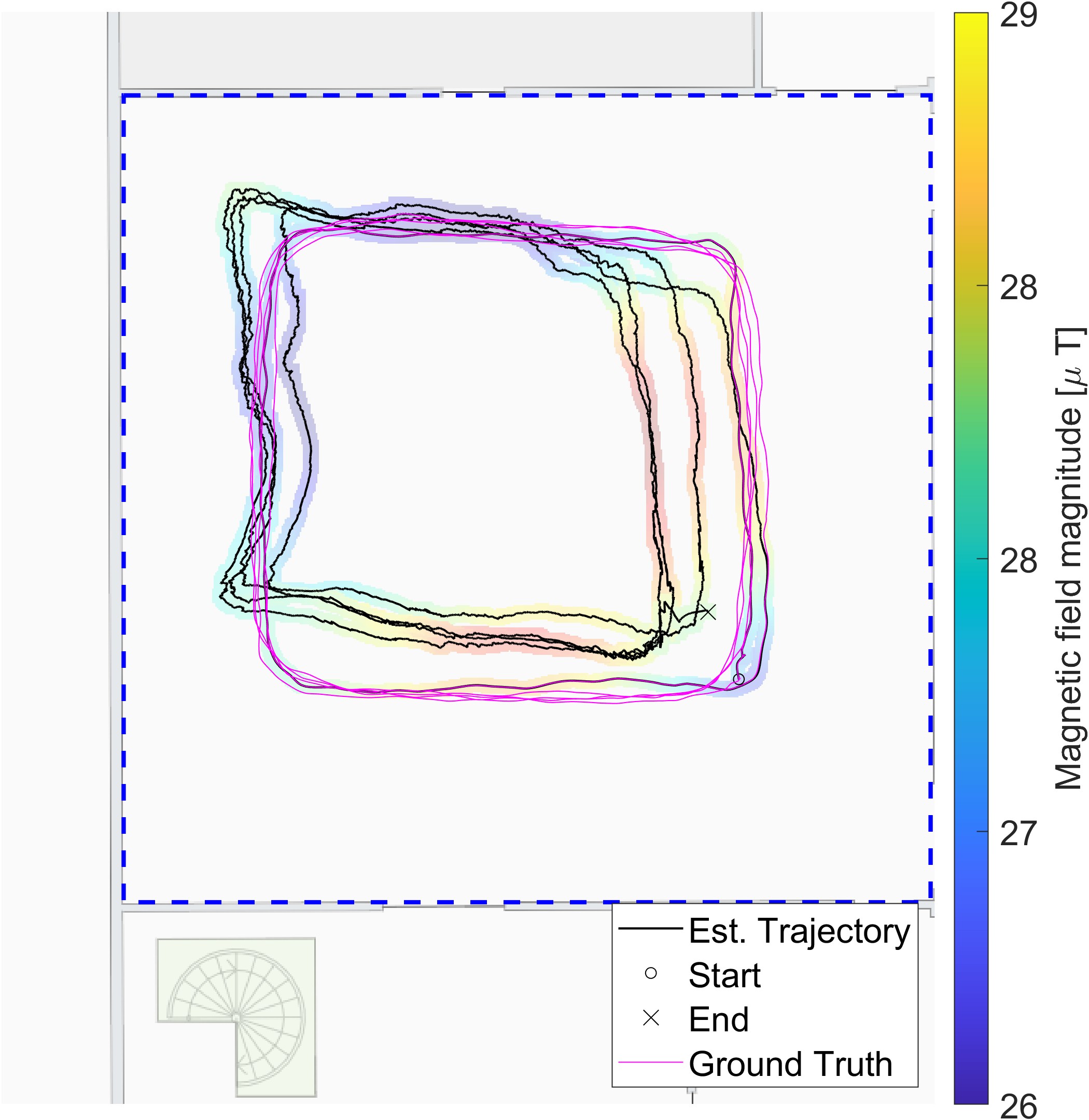}%
    }
\hfill
    \subfloat[]{%
        \includegraphics[width=0.33\textwidth]{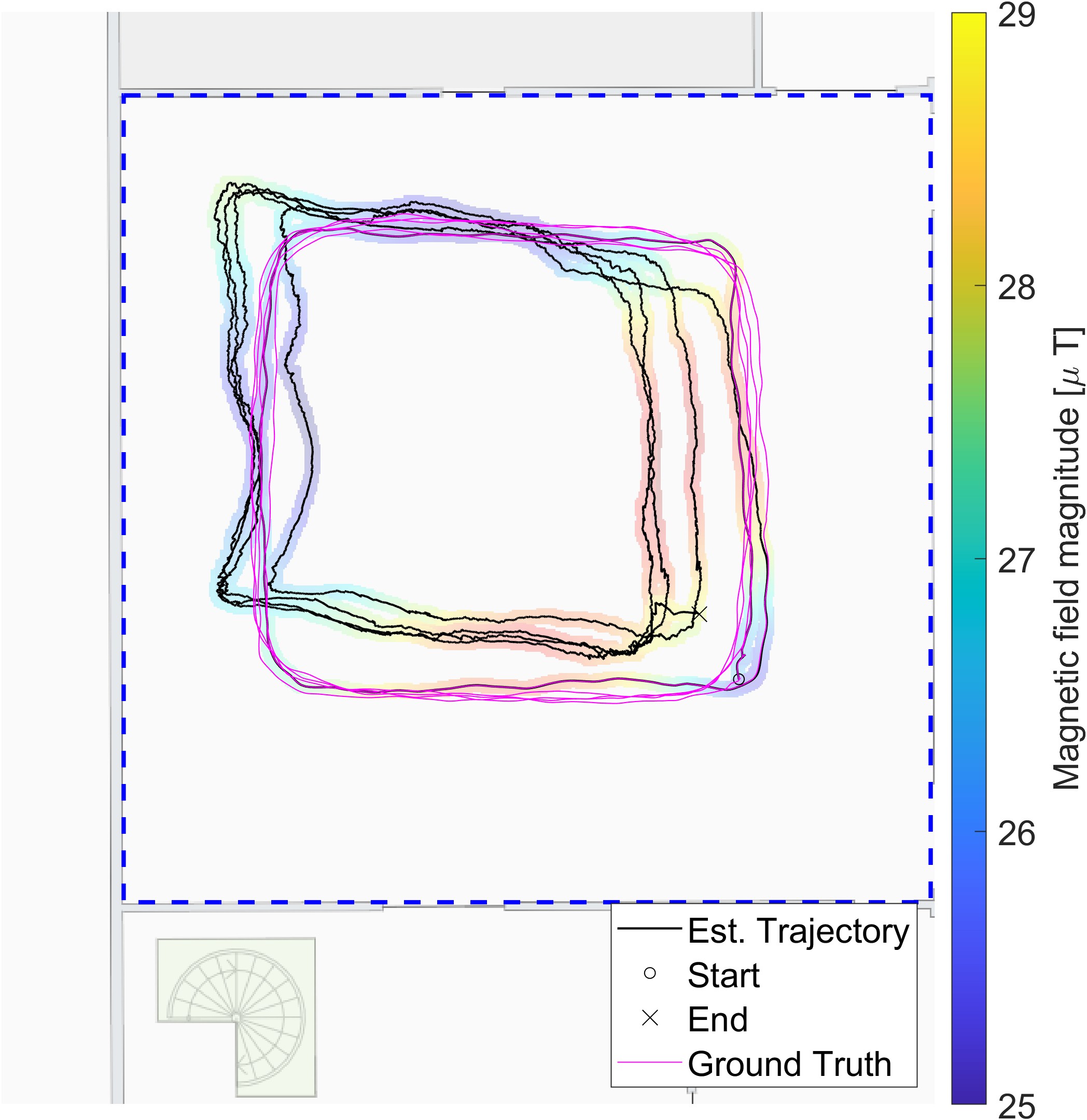}%
    }
\caption{Estimated trajectories (black) estimated by (a) MAINS, (b) the loosely coupled MI-SLAM system, and (c) the tightly coupled MI-SLAM system. The room outlined by the blue rectangular frame has dimensions of $12\text{m}\times12\text{m}$. The magnetic field maps (colored patches) built by \gls{mislam} systems are also plotted in (b) and (c), respectively.}
\label{fig: square}
\end{figure*}

The trajectories estimated by the three systems are shown in \Cref{fig: square}. It can be clearly seen that \gls{mislam} systems have smaller position drifts when compared with \gls{mains}, demonstrating the effectiveness of \gls{mislam} systems. Furthermore, \Cref{T: comparison square} gives a quantitative comparison in terms of root mean square error and computation time. As expected, \gls{mislam} systems have smaller horizontal and vertical RMSEs when compared to MAINS, because the magnetic field maps maintained in the \gls{mislam} systems allow them to correct position drift. On the other hand, they are more computationally expensive, as indicated by the computation time, especially for the tightly coupled \gls{mislam} system. Furthermore, the loosely coupled \gls{mislam} system has comparable RMSEs to those of the tightly coupled \gls{mislam} system, while being much faster.

\begin{table}[tb!]
\centering
\caption{Comparison of horizontal-vertical root mean square error (RMSE) and computation time for the three methods.}
\label{tab:method_comparison}
\begin{threeparttable}
\begin{tabular}{lccc}
\toprule
\textbf{Metric} & {\textbf{MAINS}} & {\textbf{\gls{mislam} (L.)}\tnote{*}} & {\textbf{\gls{mislam} (T.)}\tnote{$\dagger$}} \\
\midrule
Horizontal RMSE (m) & 2.64 & 1.02 & 1.05 \\
Vertical RMSE (m)   & 0.44 & 0.26 & 0.18 \\
Comp. Time (s)      & 2.1 & 2.4 & 7.1 \\
\bottomrule
\end{tabular}
  \label{T: comparison square}
  \smallskip
\scriptsize
*: loosely coupled \gls{mislam} system   $\dagger$: tightly coupled \gls{mislam} system
\end{threeparttable}
\end{table}

\subsection{More Realistic Experiments}
In these experiments, we conducted experiments over larger areas. In each experiment, a person carried the sensor array and started moving in a room equipped with a motion capture system for 20 to 40 seconds. The person then walked out of the room and explored the surrounding area for several minutes before returning to the room. In total, three types of trajectories were recorded: a trajectory with square loops, a trajectory with long straight segments, and a trajectory with spiral upward segments (climbing a spiral staircase). They are named Corridor (C), Long Corridor (LC), and Spiral Staircase (SS). The characteristics of these datasets are summarized in~\Cref{T: dataset info}.

The setup for evaluation remains largely the same as in~\Cref{subsec: proof-of-concept}, except that the number of basis functions was set to 100, 400, and 400 for datasets C, LC, and SS, respectively. Moreover, the tracked poses when the person remained in the room at the beginning of each experiment were provided to all three systems to stabilize the IMU bias estimates. 
\begin{table*}[tb!]
  \centering
\begin{threeparttable}
  \caption{Information about the datasets}
  \begin{tabular}{lllllll}
  \hline
  \hline
   Data sequence & \textbf{C-1} & \textbf{C-2} & \textbf{LC-1} & \textbf{LC-2} & \textbf{SS-1} & \textbf{SS-2}\\
    \hline
Trajectory duration\tnote{*} (s)  & 107 & 118 & 237 & 234.5 & 192 & 181\\
Trajectory duration\tnote{$\dagger$} (s)  & 72.00 & 78.12 & 205.96 & 198.18 & 141.47 & 131.83\\
Average height (m)  & 1.04 & 1.02 & 0.93 & 0.80 & 2.29 & 2.39\\
Max. height difference (m)  & 1.76 & 1.85 & 1.68 & 1.59 & 4.53 & 4.36\\
Max. magnetic field strength difference\tnote{+} ($\mu$T)& 13.59 & 14.21 & 19.51 & 19.16 & 64.28 & 45.80\\
Std. of magnetic field strength ($\mu$T)& 8.36 & 3.12 & 3.40 & 3.28 & 4.55 & 7.47\\
  \hline
  \hline
  \end{tabular}
\smallskip
\scriptsize
* including the initial part of the trajectory in the motion capture area, $\dagger$ excluding the initial part of the trajectory in the motion capture area. \\
+ the difference between the maximum and minimum magnetic field strength measured by the magnetometer array during the trajectory. \\
C: corridor  \; LC: long corridor \;     SS: spiral staircase.
  \label{T: dataset info}
   \end{threeparttable}
\end{table*}

\begin{figure*}[tb!]
    \centering
    \subfloat[]{%
        \includegraphics[width=0.27\textwidth]{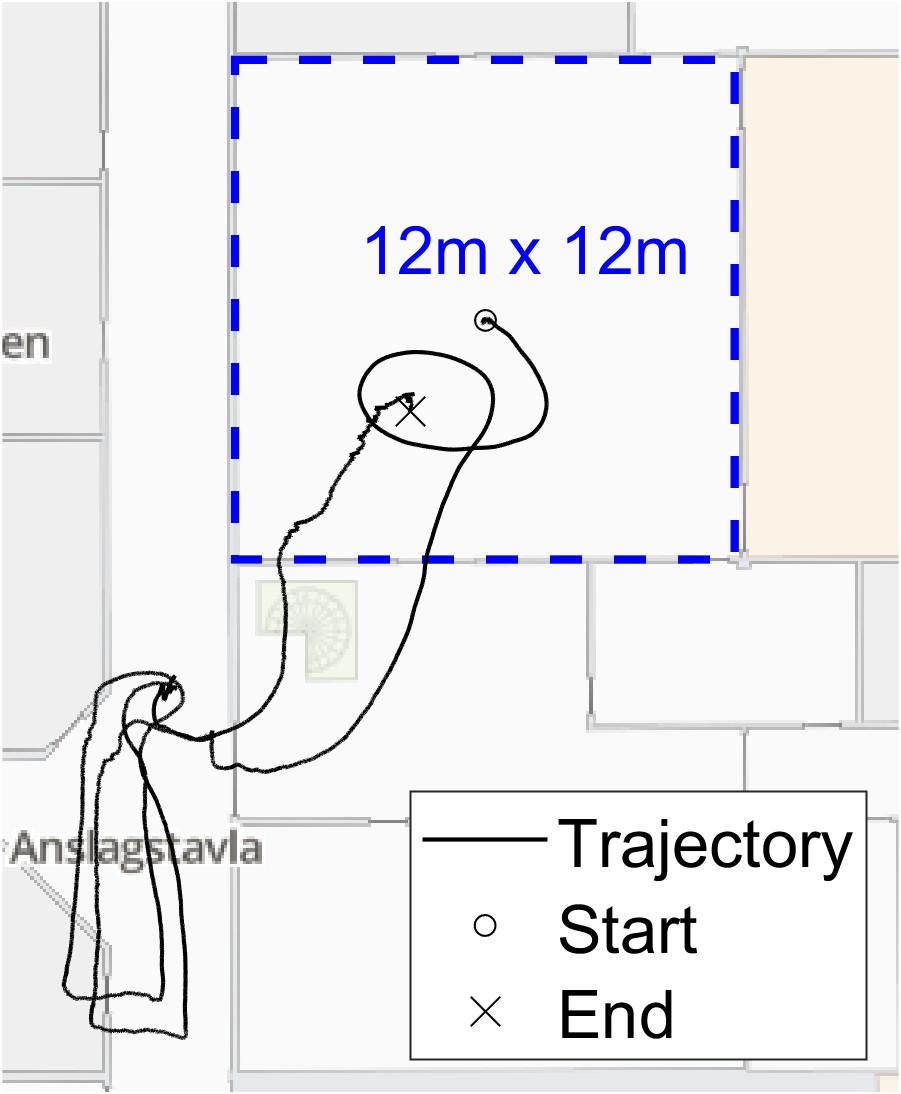}%
    }
\hfill
    \subfloat[]{%
        \includegraphics[width=0.33\textwidth]{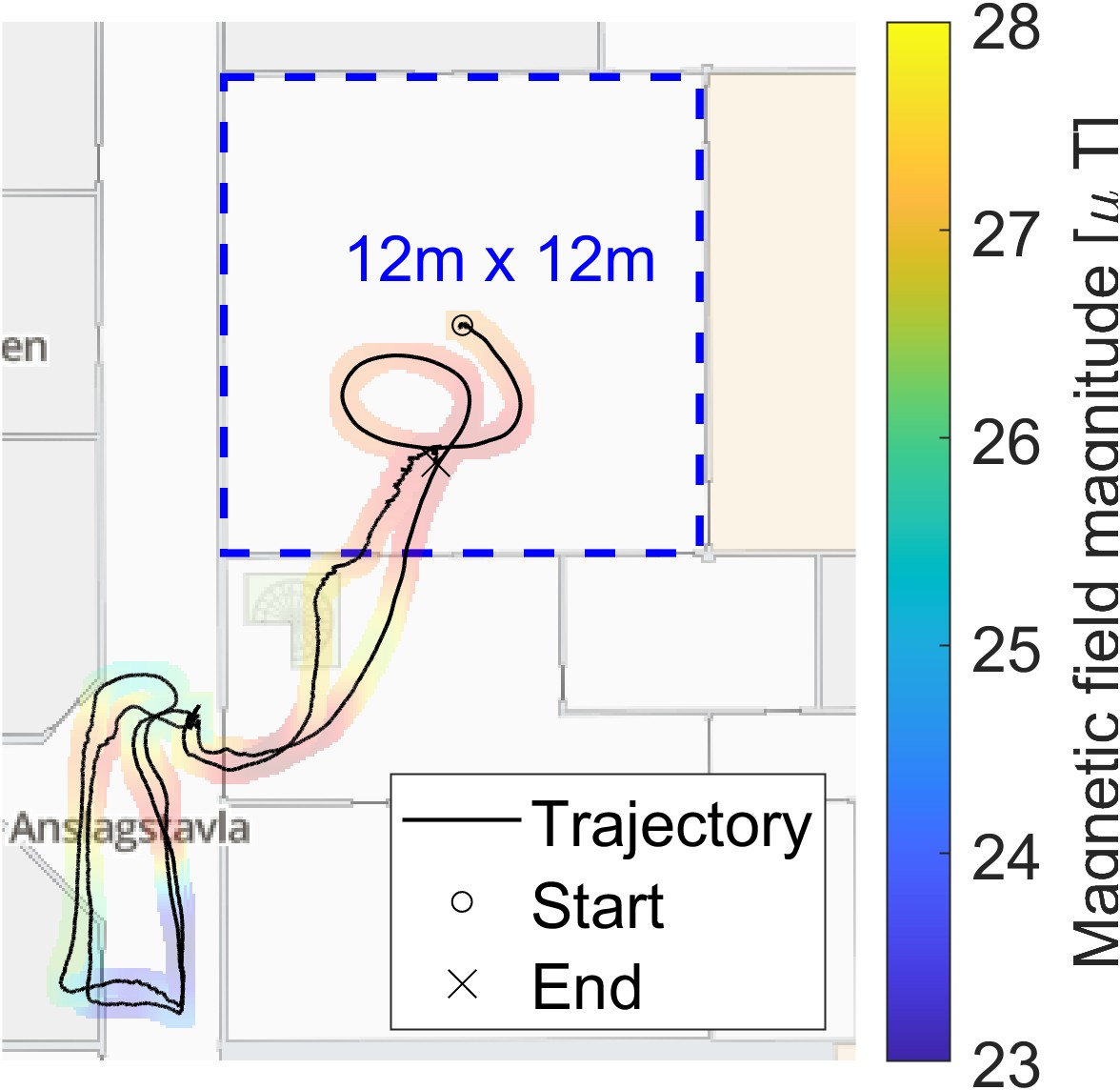}%
    }
\hfill
    \subfloat[]{%
        \includegraphics[width=0.33\textwidth]{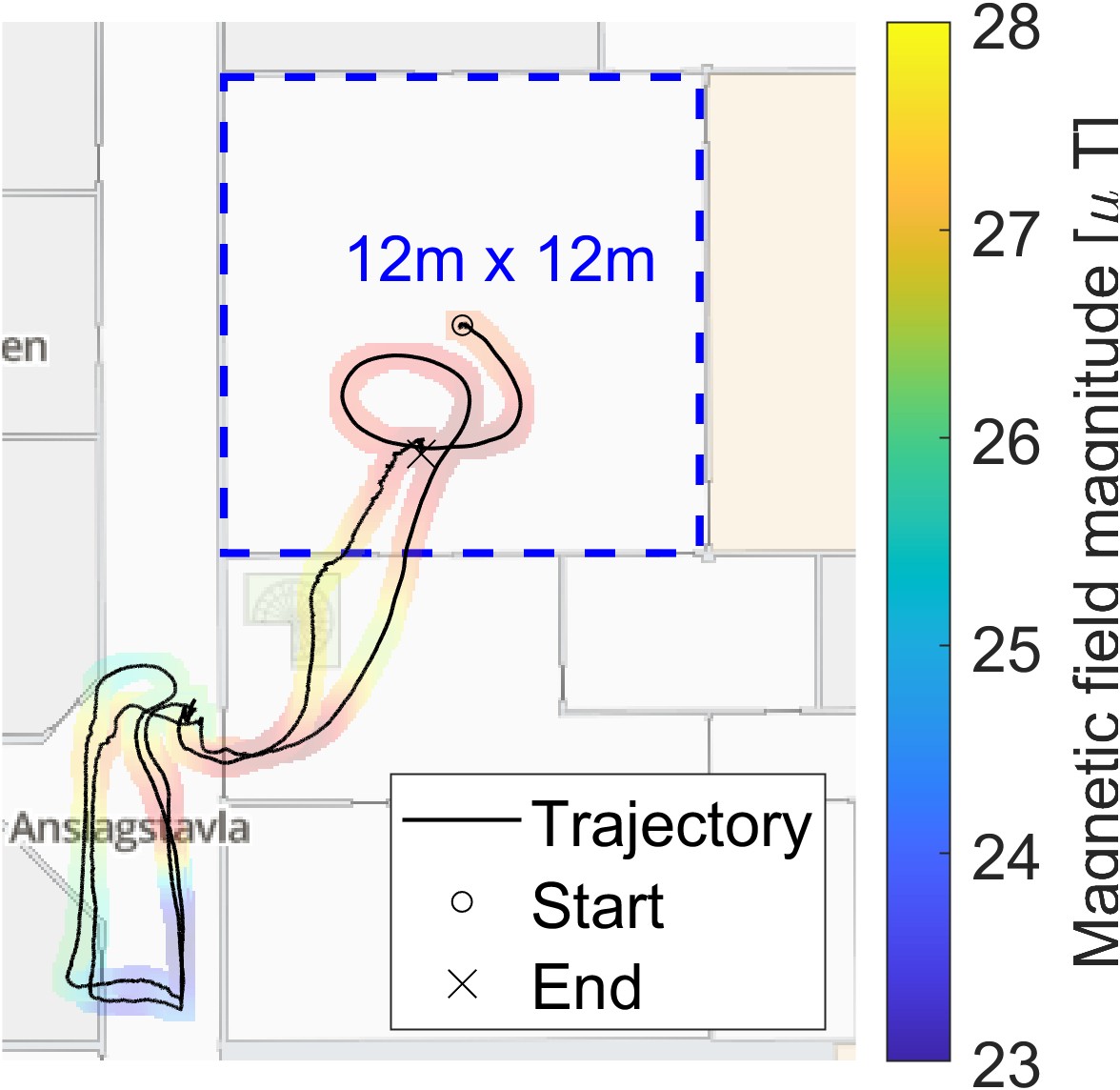}%
    }
\caption{Trajectories (C-2) estimated by (a) MAINS, (b) the loosely coupled MI-SLAM system, and (c) the tightly coupled MI-SLAM system. The room where the motion capture system is located is marked with a blue rectangular frame.}
\label{fig: corridor}
\end{figure*}

\begin{figure*}[tb!]
    \centering
    \subfloat[]{%
        \includegraphics[width=0.26\textwidth]{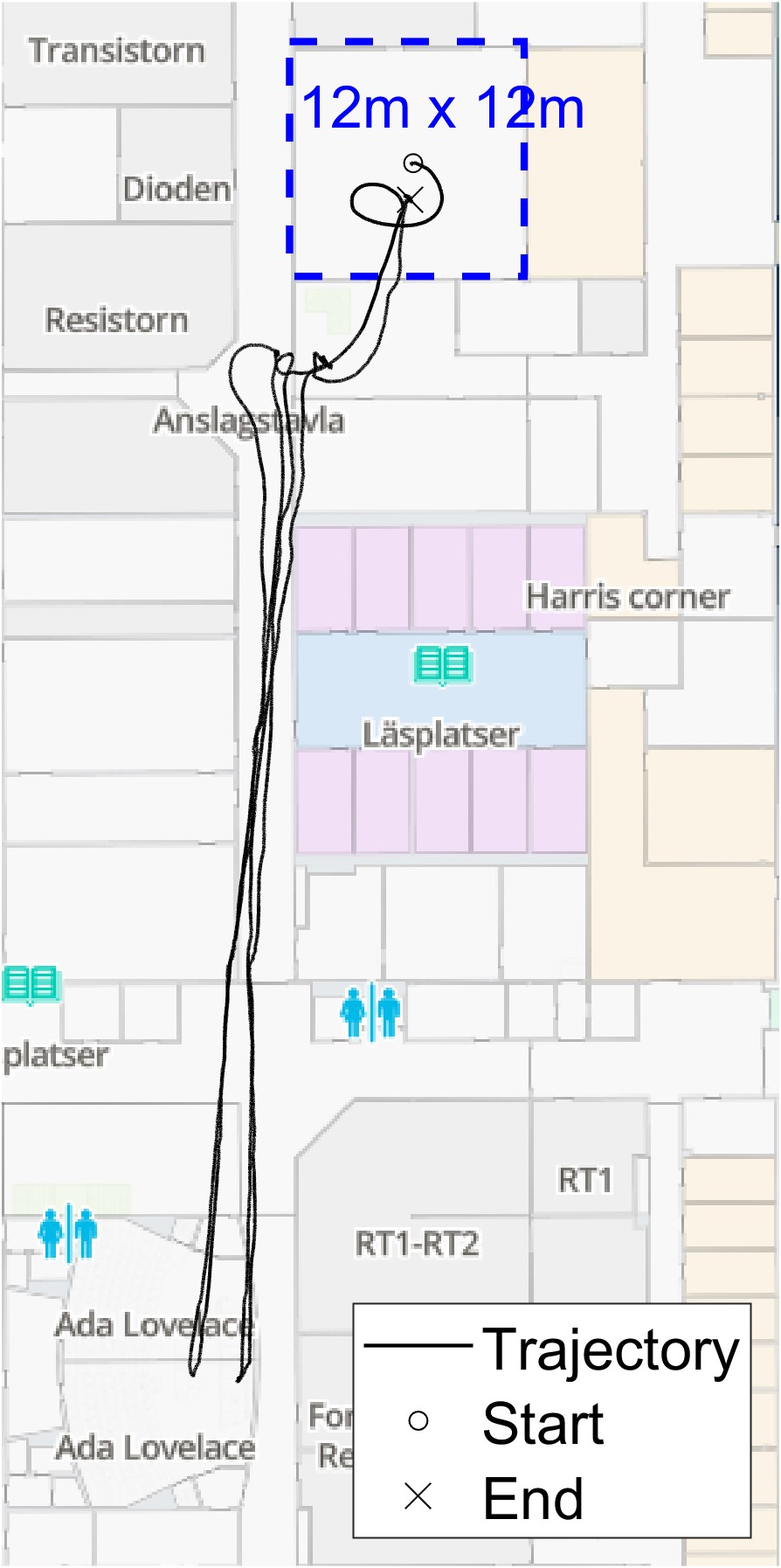}%
    }
\hfill
    \subfloat[]{%
        \includegraphics[width=0.33\textwidth]{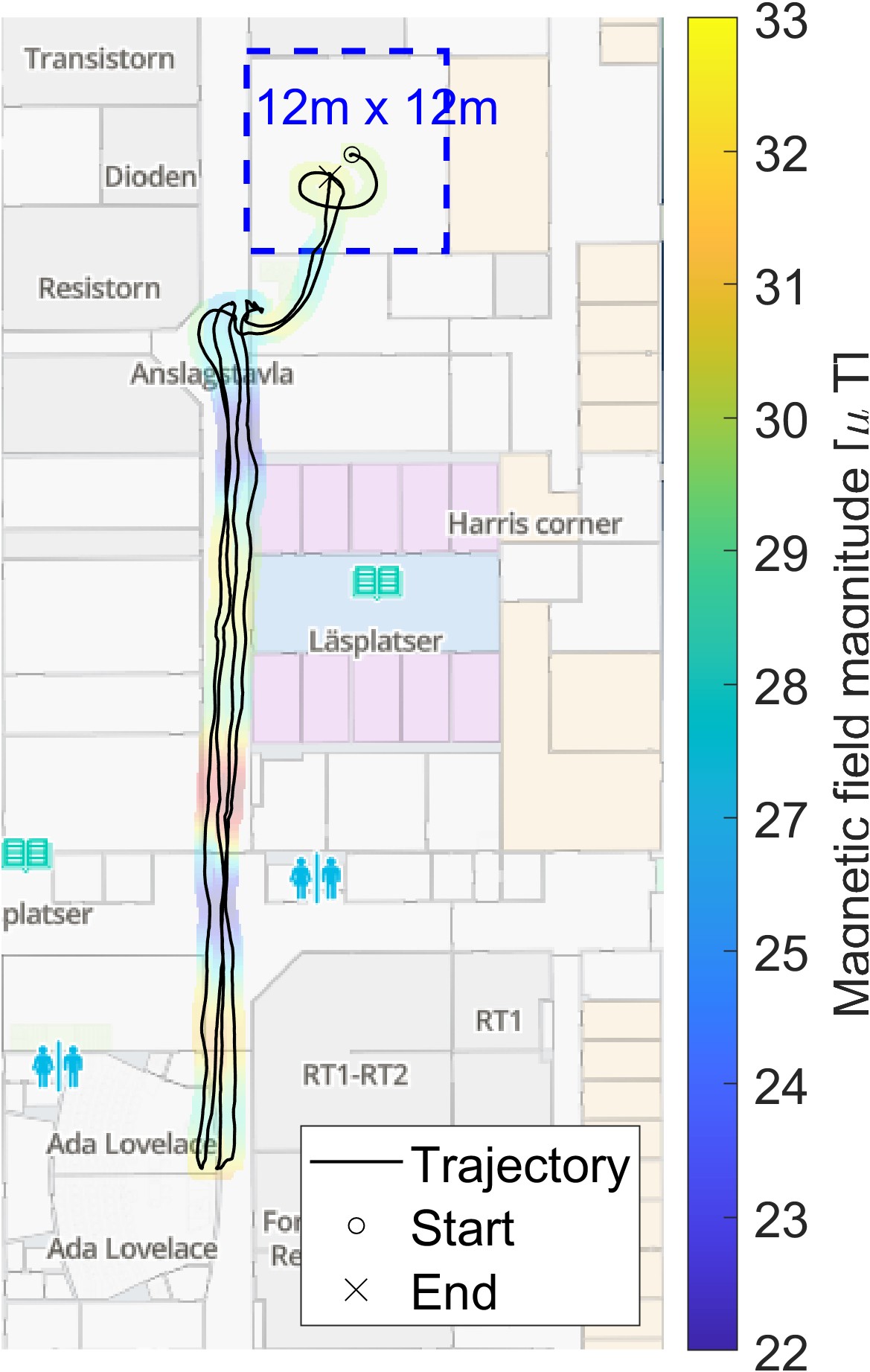}%
    }
\hfill
    \subfloat[]{%
        \includegraphics[width=0.33\textwidth]{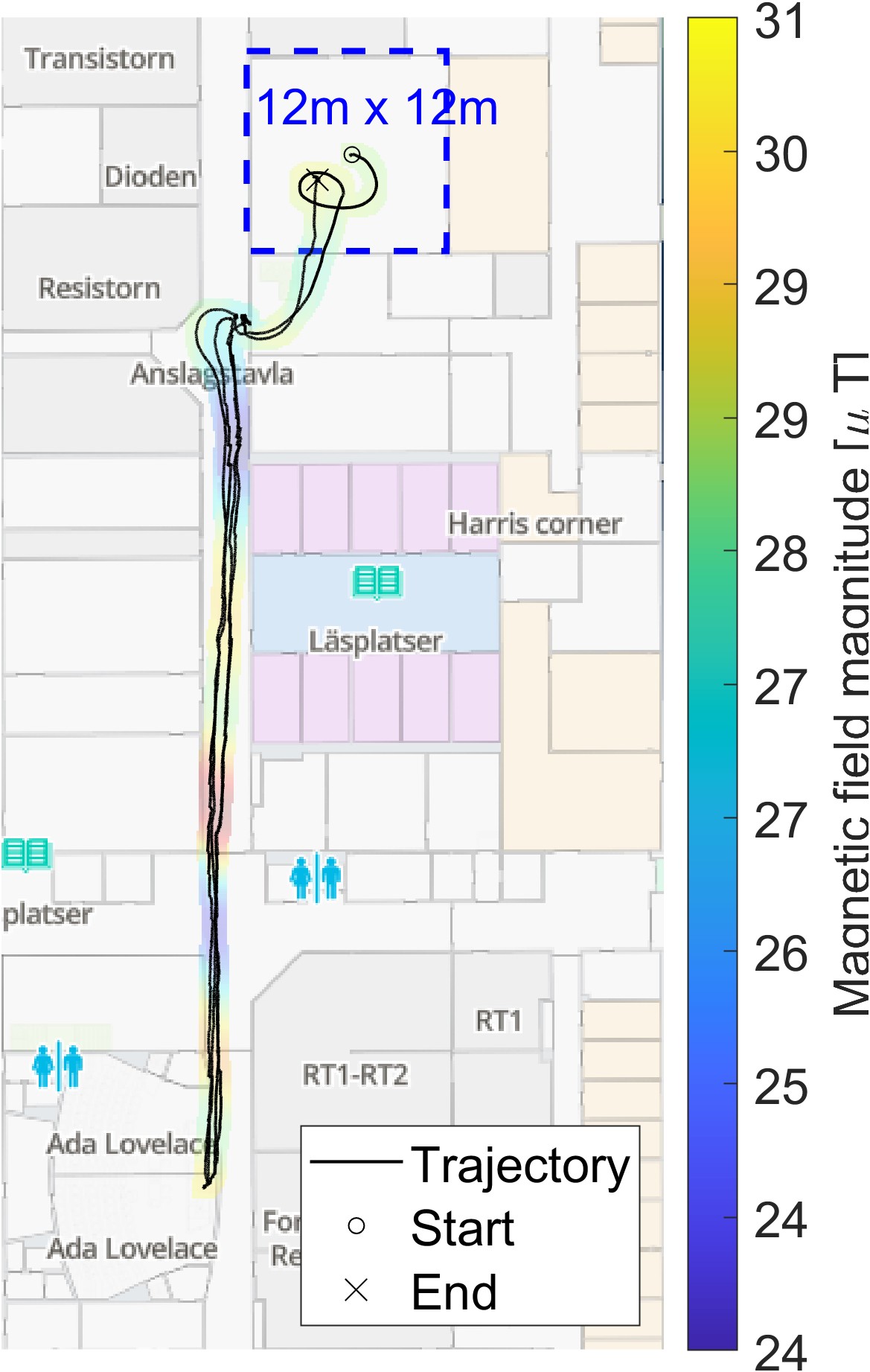}%
    }
\caption{Trajectories (LC-1) estimated by (a) MAINS, (b) the loosely coupled MI-SLAM system, and (c) the tightly coupled MI-SLAM system. The room where the motion capture system is located is marked with a blue rectangular frame.}
\label{fig: long corridor}
\end{figure*}

\begin{figure*}[tb!]
    \centering
    \subfloat[]{%
        \includegraphics[width=0.27\textwidth]{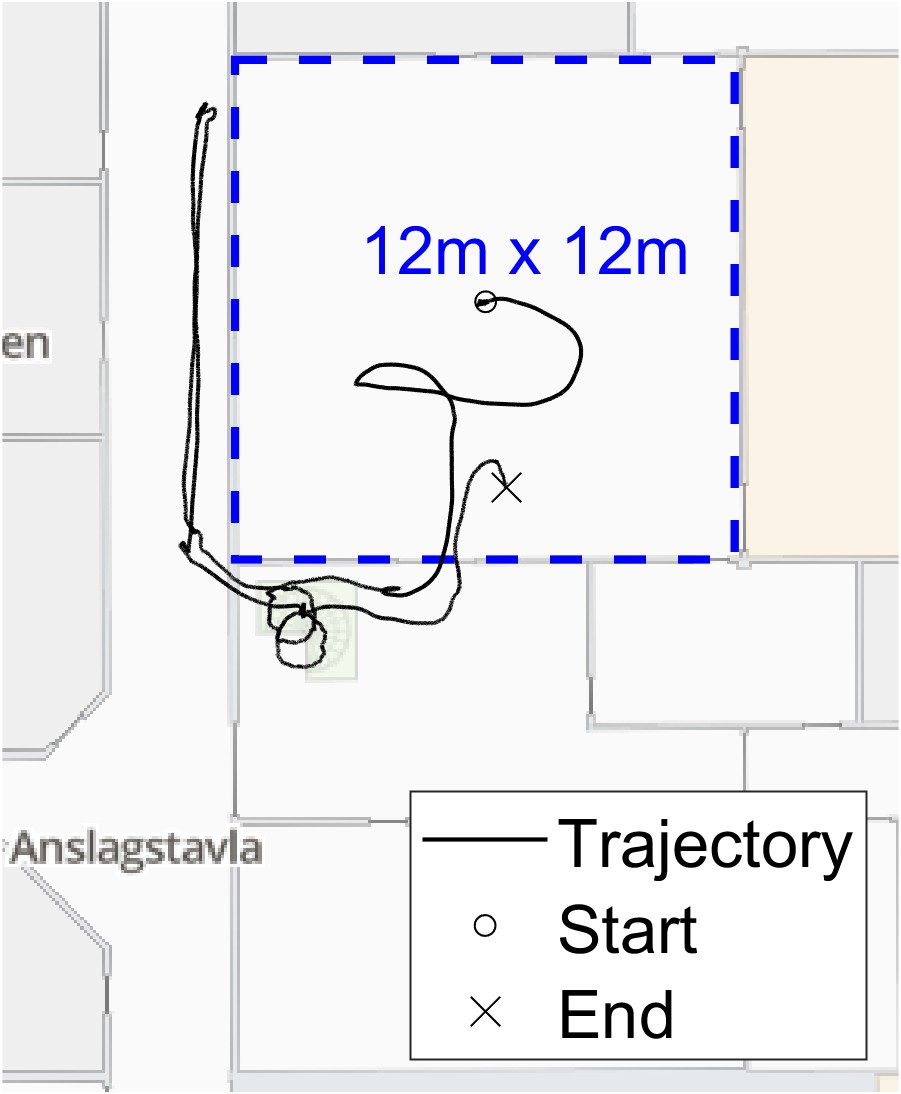}%
    }
\hfill
    \subfloat[]{%
        \includegraphics[width=0.33\textwidth]{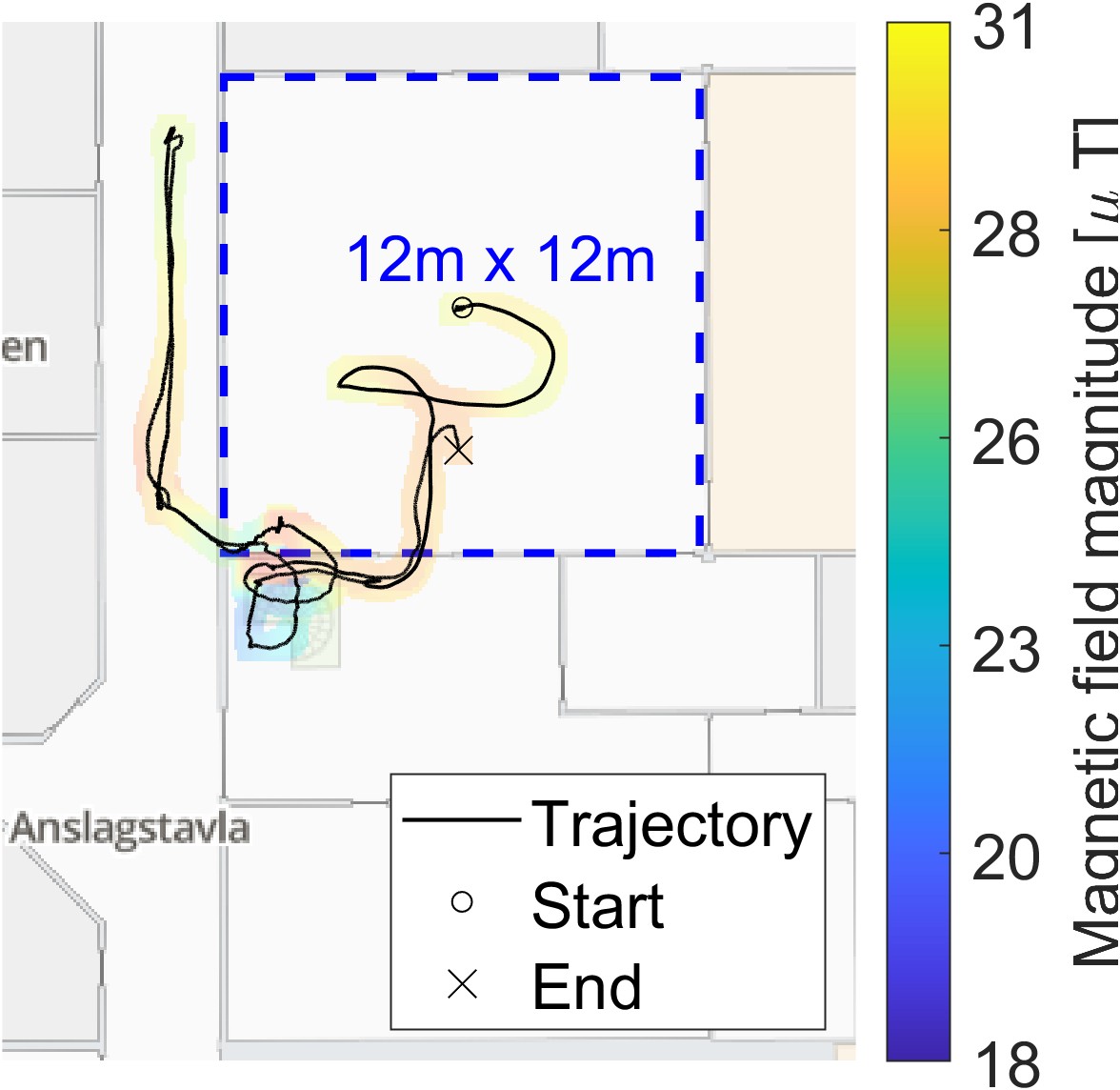}%
    }
\hfill
    \subfloat[]{%
        \includegraphics[width=0.33\textwidth]{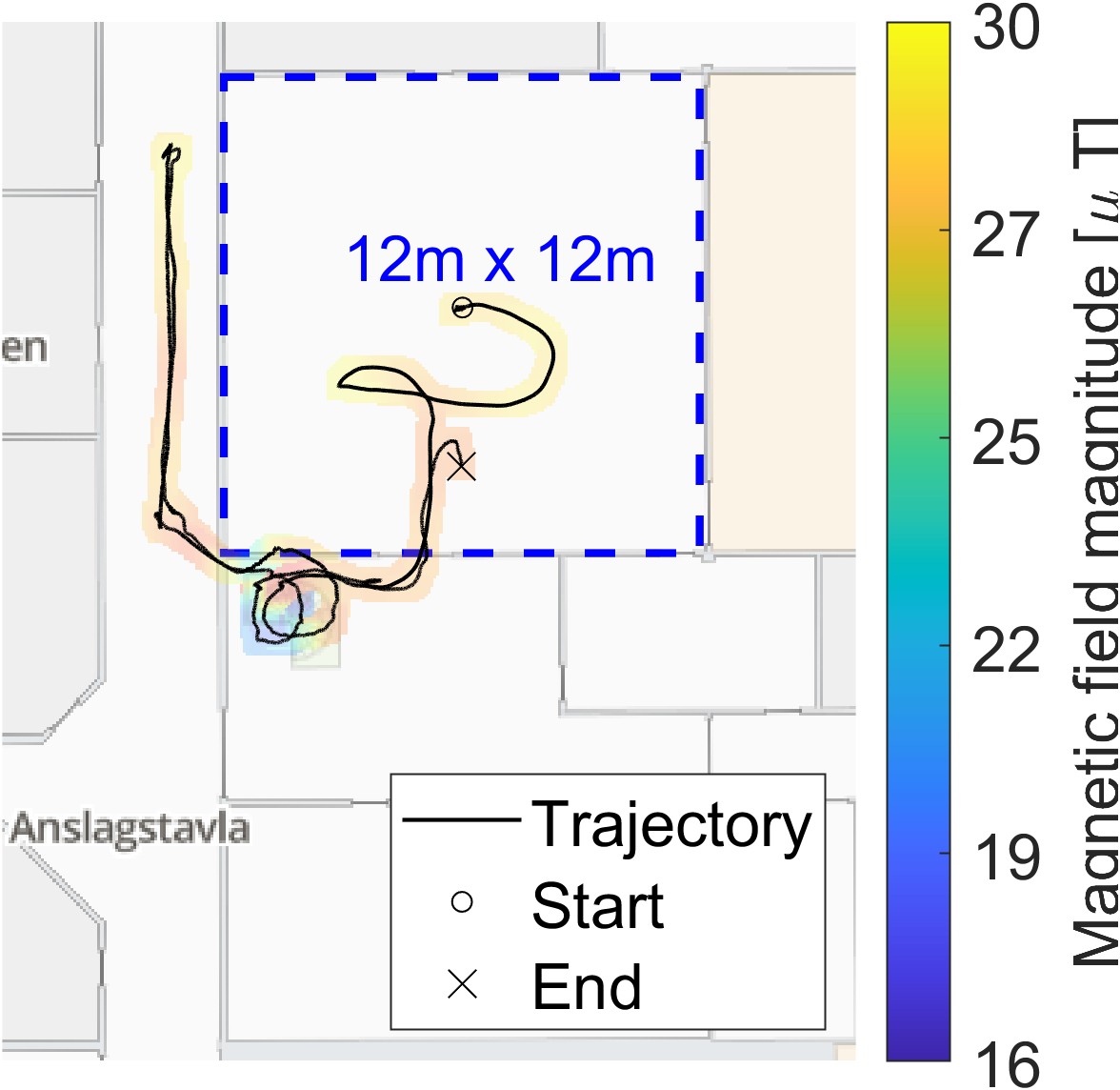}%
    }
\caption{Trajectories (SS-2) estimated by (a) MAINS, (b) the loosely coupled MI-SLAM system, and (c) the tightly coupled MI-SLAM system. The room where the motion capture system is located is marked with a blue rectangular frame.}
\label{fig: spiral staircase}
\end{figure*}

Three representative estimated trajectories are shown in~\Cref{fig: long corridor,fig: corridor,fig: spiral staircase}. Because a large portion of the trajectories were outside the motion capture area, only a qualitative evaluation of this portion of the estimated trajectories can be done. Overall, all systems are able to recover the general shape of the trajectories. The noticeable differences between the \gls{mislam} systems and MAINS can be observed in the datasets LC-2 and C-3. In these datasets, the \gls{mislam} systems produce more consistent trajectories, whereas MAINS exhibits noticeable heading or translational drift. Lastly, when comparing Fig.\ref{fig: long corridor}~(b) and~(c), it can be seen that the tightly coupled \gls{mislam} gives more concentrated straight-line segments while the loosely coupled \gls{mislam} gives more spread ones. Given that similar magnetic field magnitudes were observed on the four long straight-line segments in~Fig.\ref{fig: long corridor}~(b), the true trajectory is likely to have overlapping segments, which indicates the tightly coupled \gls{mislam} system may have performed better.

\begin{figure}[tb!]
    \centering
    \includegraphics[width=\columnwidth]{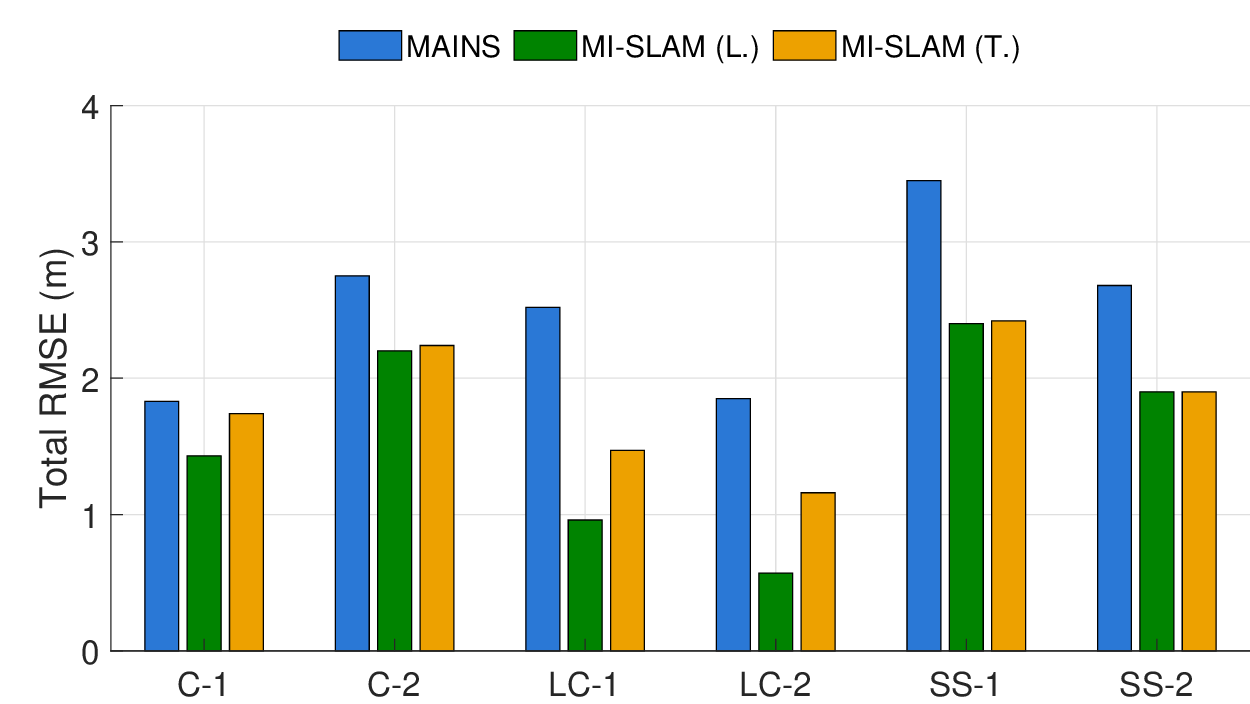}
    \caption{The total RMSE (horizontal and vertical combined) of MAINS, MI-SLAM (L.), and MI-SLAM (T.) over the final segment of the trajectory, when the person returns to the room equipped with a motion capture system.}
    \label{fig: total RMSE}
\end{figure}

\begin{figure}[tb!]
    \centering
    \includegraphics[width=\columnwidth]{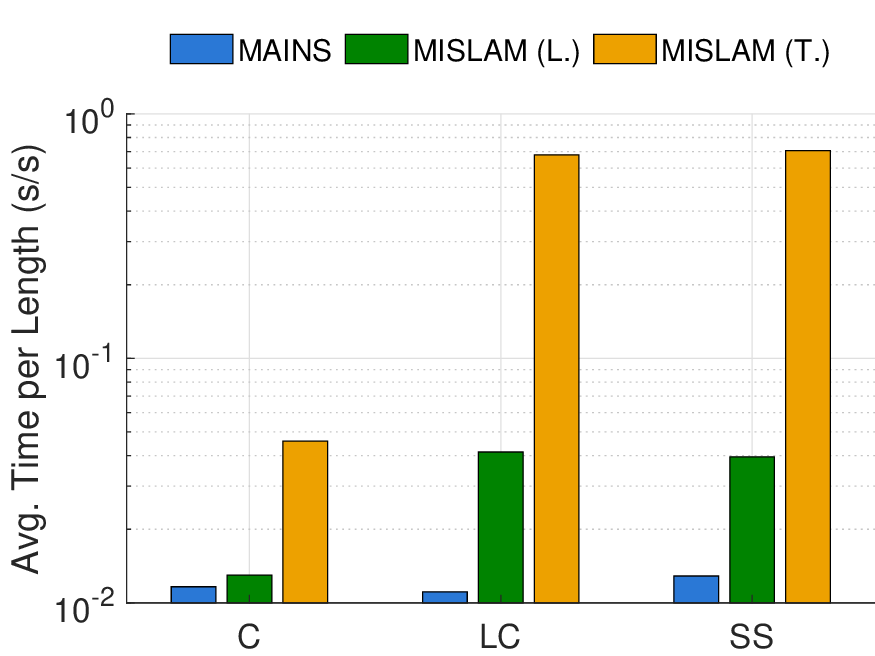}
    \caption{The averaged computation time per data length of MAINS, MI-SLAM (L.), and MI-SLAM (T.) on different datasets. The numbers of basis functions used in datasets C, LC, and SS are 100, 400, and 400, respectively. For \gls{mislam} systems, the most important factors influencing computation time are the number of basis functions and the system architecture.}
    \label{fig: avr. time}
\end{figure}

The quantitative results of the final part of the trajectories when the person returns to the room are summarized in Table~\ref{Tab: Horizontal and Vertical Error} and \Cref{fig: total RMSE,fig: avr. time}. Table~\ref{Tab: Horizontal and Vertical Error} reports the horizontal and vertical RMSE. As shown in the table, the horizontal errors of all systems on datasets SS-1 and SS-2 are considerably larger than those on the other datasets, which is likely due to the vertical motion associated with ascending and descending the spiral staircase. Overall, MI-SLAM systems achieve smaller horizontal and vertical errors than MAINS on most datasets. The corresponding total RMSEs over the same trajectory segment are shown in Fig.~\ref{fig: total RMSE}. On this evaluation segment, both \gls{mislam} systems consistently yield lower total RMSE than MAINS across all datasets. Among the three systems, the loosely coupled \gls{mislam} system achieves the lowest total RMSE on this segment, although this observation is specific to the final portion of the trajectories considered in the evaluation. Finally, Fig.~\ref{fig: avr. time} compares the computational efficiency of the three systems in terms of the average computation time per unit data length. As expected, MAINS is the fastest system because it performs magnetic-inertial odometry without maintaining a globally consistent magnetic field map. The loosely coupled \gls{mislam} system requires more computation than MAINS but remains substantially more efficient than the tightly coupled \gls{mislam} system. For both MI-SLAM systems, the computation time increases with the number of basis functions, indicating that the map complexity and system architecture are the primary factors affecting computational cost.

To sum up, \gls{mislam} systems have lower total RMSEs on the final part of the trajectory than \gls{mains}. Moreover, the loosely coupled \gls{mislam} system is more computationally efficient and slightly more accurate on the final part of the trajectory than the tightly coupled \gls{mislam} system, although Fig.~\ref{fig: long corridor} shows that the tightly coupled \gls{mislam} system may be more accurate on the major part of the trajectory outside the motion capture room. These results indicate that the system architecture has the greatest influence on the computational cost of the proposed systems.


\begin{table}[tb!]
\centering
\begin{threeparttable}
\caption{Horizontal and Vertical RMSE of the Final Segment of the Trajectories (meters)\\}
\begin{tabular}{lccc}
\toprule
 & MAINS  & MI-SLAM (L.)  & MI-SLAM (T.)\\
\midrule
C-1 & 1.50 \textbar\ 0.33 & 1.08 \textbar\ 0.35 & 1.28 \textbar\ 0.46 \\
C-2 & 2.13 \textbar\ 0.62 & 2.09 \textbar\ 0.11 & 2.14 \textbar\ 0.10 \\
LC-1 & 1.41 \textbar\ 1.11 & 0.75 \textbar\ 0.21 & 1.30 \textbar\ 0.17 \\
LC-2 & 0.51 \textbar\ 1.34 & 0.46 \textbar\ 0.11 & 0.95 \textbar\ 0.21 \\

SS-1 & 2.92 \textbar\ 0.53 & 1.90 \textbar\ 0.50 & 2.28 \textbar\ 0.14 \\
SS-2 & 2.42 \textbar\ 0.26 & 1.53 \textbar\ 0.37 & 1.86 \textbar\ 0.04 \\
\bottomrule
\end{tabular}
\label{Tab: Horizontal and Vertical Error}
\end{threeparttable}
\end{table}

\subsection{Ablation Study on the Effect of Height Measurements and Inertial Sensor Quality}
\begin{figure*}[tb!]
    \centering
    \subfloat[Dataset: C]{%
        \includegraphics[width=0.31\textwidth]{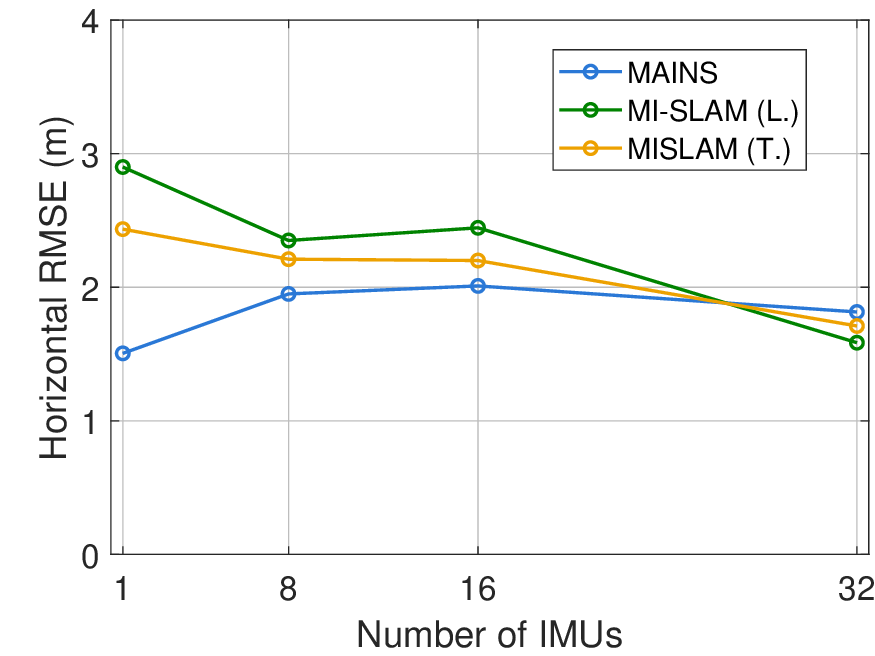}
        \label{fig:horizontal_error_LC}
    }\hfill
    \subfloat[Dataset: LC]{%
        \includegraphics[width=0.31\textwidth]{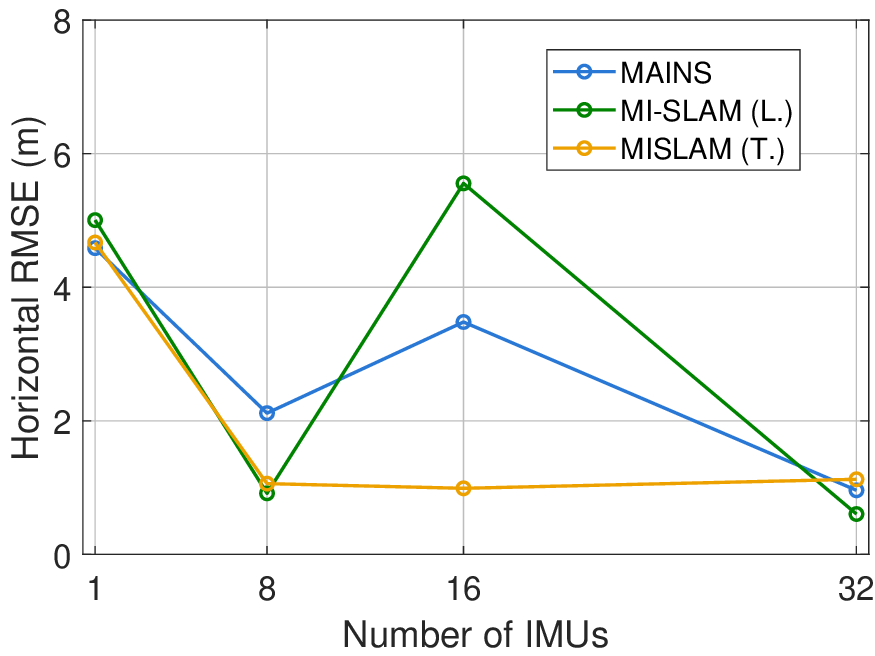}
        \label{fig:horizontal_error_C}
    }\hfill
    \subfloat[Dataset: SS]{%
        \includegraphics[width=0.31\textwidth]{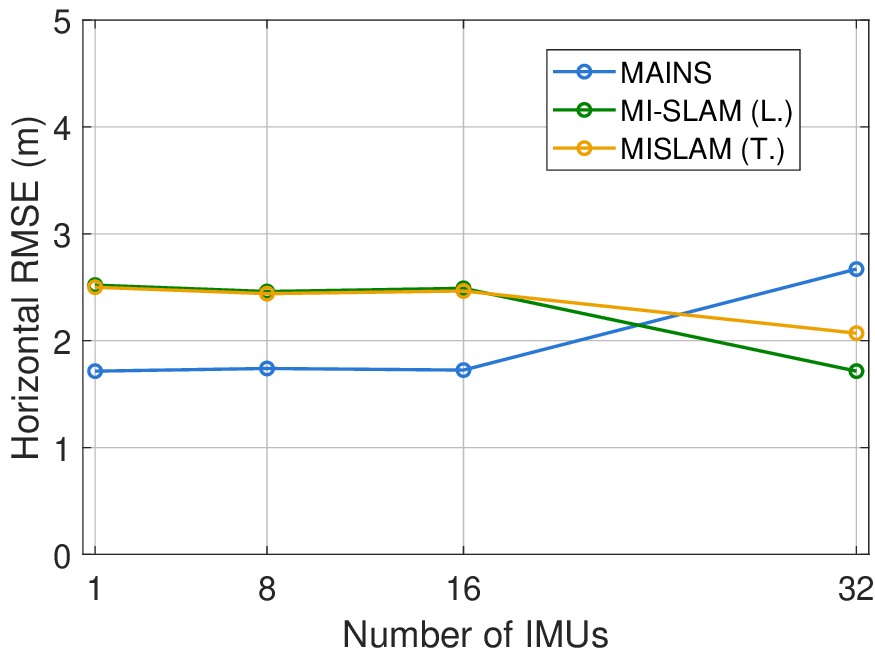}
        \label{fig:horizontal_error_SS}
    }
    \caption{Horizontal RMSEs using IMU measurements of different qualities. The ~\href{https://www.inertialelements.com/osmium-mimu4444.html}{Osmium MIMU 4844 IMU} is a 32-IMU array; its measurement quality is varied by averaging different numbers of IMUs in the array.}
    \label{fig:horizontal_error_all_three}
\end{figure*}

In these experiments, the effect of height measurements and inertial sensor quality on the performance of all systems was investigated.
First, noisy height measurements (standard deviation of 25 cm) given by a barometer were provided for all three systems, and the results are summarized in TABLE~\ref{Tab: Horizontal and Vertical Error with Baro.}. Compared with TABLE~\ref{Tab: Horizontal and Vertical Error}, the vertical errors of all three systems get smaller, which is expected, but the horizontal errors only change slightly. This indicates that MAINS and MI-SLAM systems are not heavily reliant on the height measurements. However, the height measurements are expected to have a more significant impact if explicit loop closure detection is to be performed in the SLAM systems, since they can help distinguish different floors in a multi-floor building.
For the second part, the IMU measurement quality is varied by averaging the measurements from different numbers of IMUs in the \href{https://www.inertialelements.com/osmium-mimu4444.html}{Osmium MIMU 4844}. Averaging measurements from more IMUs yields higher-quality inertial measurements. The resulting RMSEs as a function of the number of IMUs are shown in Fig.~\ref{fig:horizontal_error_all_three}. Overall, the performance of all three systems remains largely stable once the measurement quality reaches that obtained by averaging 8 IMUs, except on the LC dataset. As shown in Fig.~\ref{fig:horizontal_error_all_three}~(b), the abrupt increase in the RMSEs of MAINS and the loosely coupled \gls{mislam} system when averaging 16 IMUs is likely attributable to experimental randomness. 

\begin{table}[tb!]
\centering
\begin{threeparttable}
\caption{Horizontal and Vertical Error at the End of the Trajectories Using Height Measurements (meters)}
\begin{tabular}{lccc}
\toprule
 & MAINS  & MI-SLAM (L.) & MI-SLAM (T.)\\
\midrule
C-1& 1.58 \textbar\ 0.10 & 1.12 \textbar\ 0.01 & 1.29 \textbar\ 0.13 \\
C-2& 2.13 \textbar\ 0.01 & 2.13 \textbar\ 0.04 & 2.19 \textbar\ 0.01 \\
LC-1 & 1.37 \textbar\ 0.16 & 0.88 \textbar\ 0.12 & 1.34 \textbar\ 0.17 \\
LC-2 & 0.29 \textbar\ 0.16 & 0.55 \textbar\ 0.23 & 0.95 \textbar\ 0.13 \\

SS-1& 3.42 \textbar\ 0.03 & 1.89 \textbar\ 0.05 & 2.30 \textbar\ 0.02 \\
SS-2& 2.56 \textbar\ 0.20 & 1.40 \textbar\ 0.21 & 1.92 \textbar\ 0.20 \\
\bottomrule
\end{tabular}
\label{Tab: Horizontal and Vertical Error with Baro.}
\end{threeparttable}
\end{table}

\section{Conclusion}
In this paper, a loosely coupled and a tightly coupled \gls{mislam} systems are proposed. Both systems rely solely on commonly available low-cost sensors: an inertial measurement unit and several magnetometers. Experiment results demonstrate that the proposed \gls{mislam} systems outperform the state-of-the-art magnetic-inertial odometry system, \gls{mains}, achieving a positioning RMSE below 1 meter on a trajectory lasting approximately 200 seconds. Moreover, the loosely coupled \gls{mislam} system is more computationally efficient and slightly more accurate on the final part of the trajectory than the tightly coupled \gls{mislam} system. Furthermore, ablation studies show that incorporating a barometer primarily benefits \gls{mislam} systems' performance in the vertical direction and that IMU quality can be lowered by a factor of 4 without compromising much positioning performance. These results showcase the feasibility of developing magnetic-field-based \gls{slam} systems with low-cost sensors and the potential of applying the proposed systems in emergency response scenarios such as mine or fire rescue, where one cannot rely on a GNSS or visual-based localization system.
\appendix
The formulas for evaluating the block matrices in~\eqref{eq: odometry cov} are
\begin{subequations} \label{eq: variance}
\begin{align}
     \text{var}(\Delta \hat{p}_{i,j}) &= \text{var}(\hat{p}_{j}) + \text{var}(\hat{p}_{i}) - \text{K}_{\hat{p}_{i}\hat{p}_{j}} -\text{K}_{\hat{p}_{j}\hat{p}_{i}} \label{eq: displacement var}\\
     \text{var}(\Delta \hat{q}_{i,j}) &= \text{var}(\hat{q}_{j}) + \Delta\hat{R}_{i,j}^{\top}\text{var}(\hat{q}_{i})\Delta\hat{R}_{i,j} \nonumber\\ &\quad- \Delta\hat{R}_{i,j}^{\top}\text{K}_{\hat{q}_{i}\hat{q}_{j}}- \text{K}_{\hat{q}_{j}\hat{q}_{i}} \Delta\hat{R}_{i,j} \label{eq: ori change var}\\
     \text{K}_{\Delta \hat{p}_{i,j} \Delta \hat{q}_{i,j}}&=(\text{K}_{p_j q_i}-\text{K}_{p_i q_i})\Delta\hat{R}_{i,j} - \text{K}_{p_j q_j} + \text{K}_{p_i q_j}\\
     \text{K}_{\Delta \hat{q}_{i,j}\Delta \hat{p}_{i,j} }&=\text{K}_{\Delta \hat{p}_{i,j} \Delta \hat{q}_{i,j}}^{\top}
\end{align}
\end{subequations}
Here the covariance of the quaternion estimate is defined as the covariance of the associated random error vector under the local perturbation model, i.e., $q=\hat{q}\otimes \text{Exp}(\delta q)$, where $q$ denotes the true quaternion value, $\delta q\in\mathbb{R}^3$ denotes the error vector, and $\text{Exp}(\cdot)$ denotes the operator that maps a vector in $\mathbb{R}^3$ to a quaternion, see~\cite{Joan2017Quaternion} for details. 

For simplicity, we assume that the cross-covariance between $\Delta \hat{p}_{i,j}$ and $\Delta \hat{q}_{i,j}$ is negligible, i.e., $\text{K}_{\Delta \hat{p}_{i,j} \Delta \hat{q}_{i,j}}=\text{K}_{\Delta \hat{q}_{i,j}\Delta \hat{p}_{i,j}}=0$, which is justified by the fact that $\text{K}_{p_iq_i}$ is small observed in filtering. To evaluate $\text{var}(\Delta \hat{p}_{i,j})$ and $\text{var}(\Delta \hat{q}_{i,j})$, we only need to compute two additional cross-covariance, $\text{K}_{\hat{p}_i\hat{p}_j}$ and $\text{K}_{\hat{q}_i\hat{q}_j}$ during filtering, because $\text{var}(\hat{p}_i)$ and $\text{var}(\hat{q}_i)$ on the right-hand side of~\eqref{eq: displacement var} and~\eqref{eq: ori change var} are readily available from the error-state Kalman filter. The computation process is given in~\Cref{alg: cross covariance} without proof.
\begin{algorithm}[tb!]
\caption{Compute Cross-covariance}
\label{alg: cross covariance}
\begin{algorithmic}[1]
\REQUIRE Covariance matrix $P_{i|i}$, indices $i$, $j$
\STATE $P_{i,i}^{\text{cross}} = P_{i|i}$ \COMMENT{$P_{i|i}$: posterior error state covariance at time $i$}
\FOR{$k = i$ \TO $j-1$}
    \STATE $P_{i,k+1}^{\text{cross},-}=F_k P_{i,k}^{\text{cross}}$ \COMMENT{$F_k$: error state transition matrix}
    \STATE $P_{i,k+1}^{\text{cross}}=(I-K_{k+1}H_{k+1})P_{i,k+1}^{\text{cross},-}$ \COMMENT{$K_k$: Kalman gain, $H_k$: measurement Jacobian}
\ENDFOR
\STATE Extract the position and orientation cross-covariance blocks from $P_{i,j}^{\text{cross}}$
\STATE \textbf{return} $\text{K}_{\hat{p}_{i}\hat{p}_{j}}$ and $\text{K}_{\hat{q}_{i}\hat{q}_{j}}$ 
\end{algorithmic}
\end{algorithm}

\bibliographystyle{IEEEtran}
\bibliography{ref}

\begin{IEEEbiography}[{\includegraphics[width=1in,height=1.25in,clip,keepaspectratio]{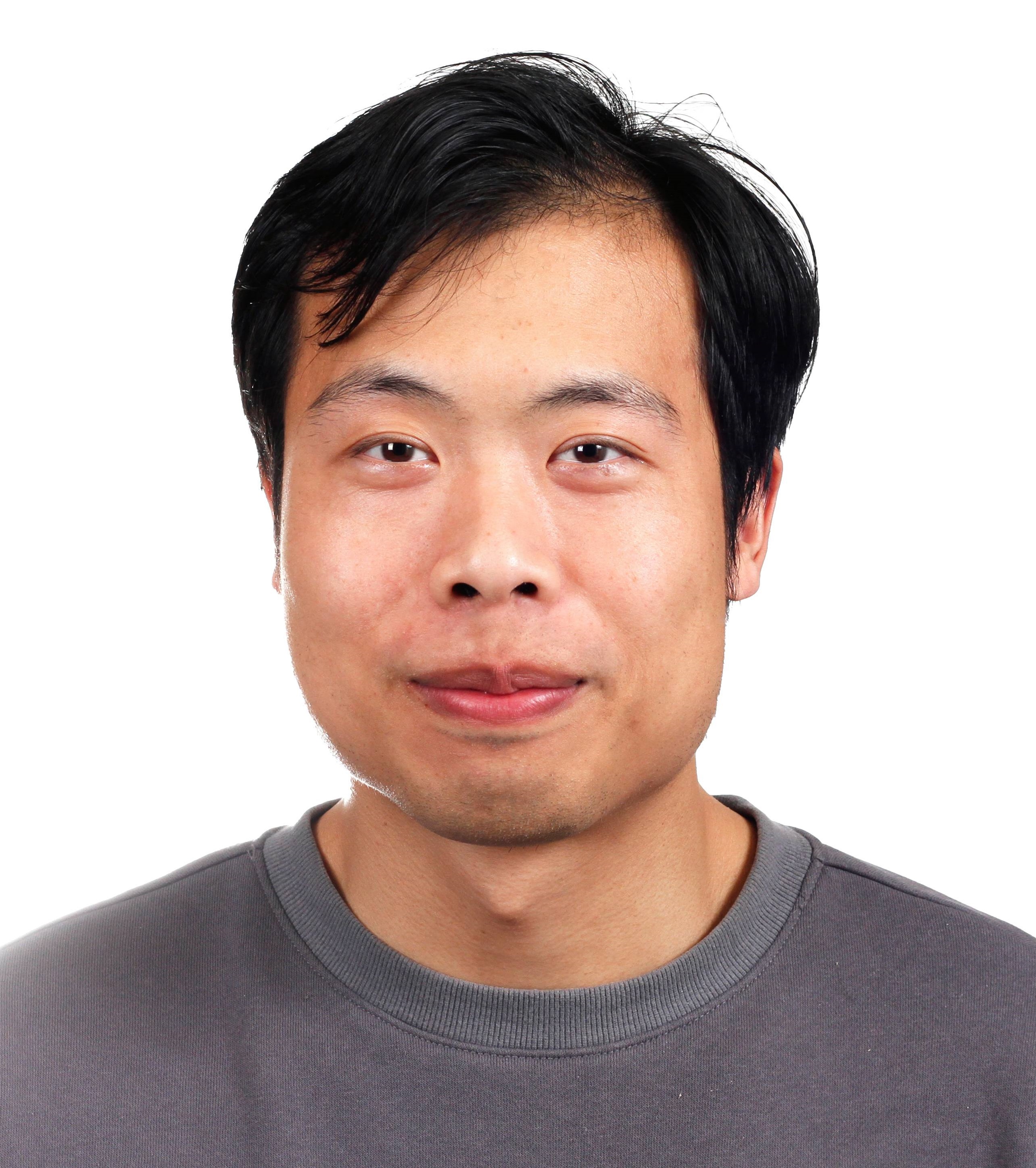}}]{Chuan Huang} (Student member, IEEE) received the B.Sc. from Beihang University in 2018 and the M.Sc. degree from China Electronics Technology Group Corporation Academy of Electronics and Information Technology in 2021. From 2021 to 2024, he studied at Linköping University, Sweden, and he is now a PhD student at the KTH Royal Institute of Technology, Stockholm, Sweden. His main research interest is magnetic field-based positioning and sensor calibration.
\end{IEEEbiography}

\begin{IEEEbiography}
[{\includegraphics[width=1in,height=1.25in,clip,keepaspectratio]{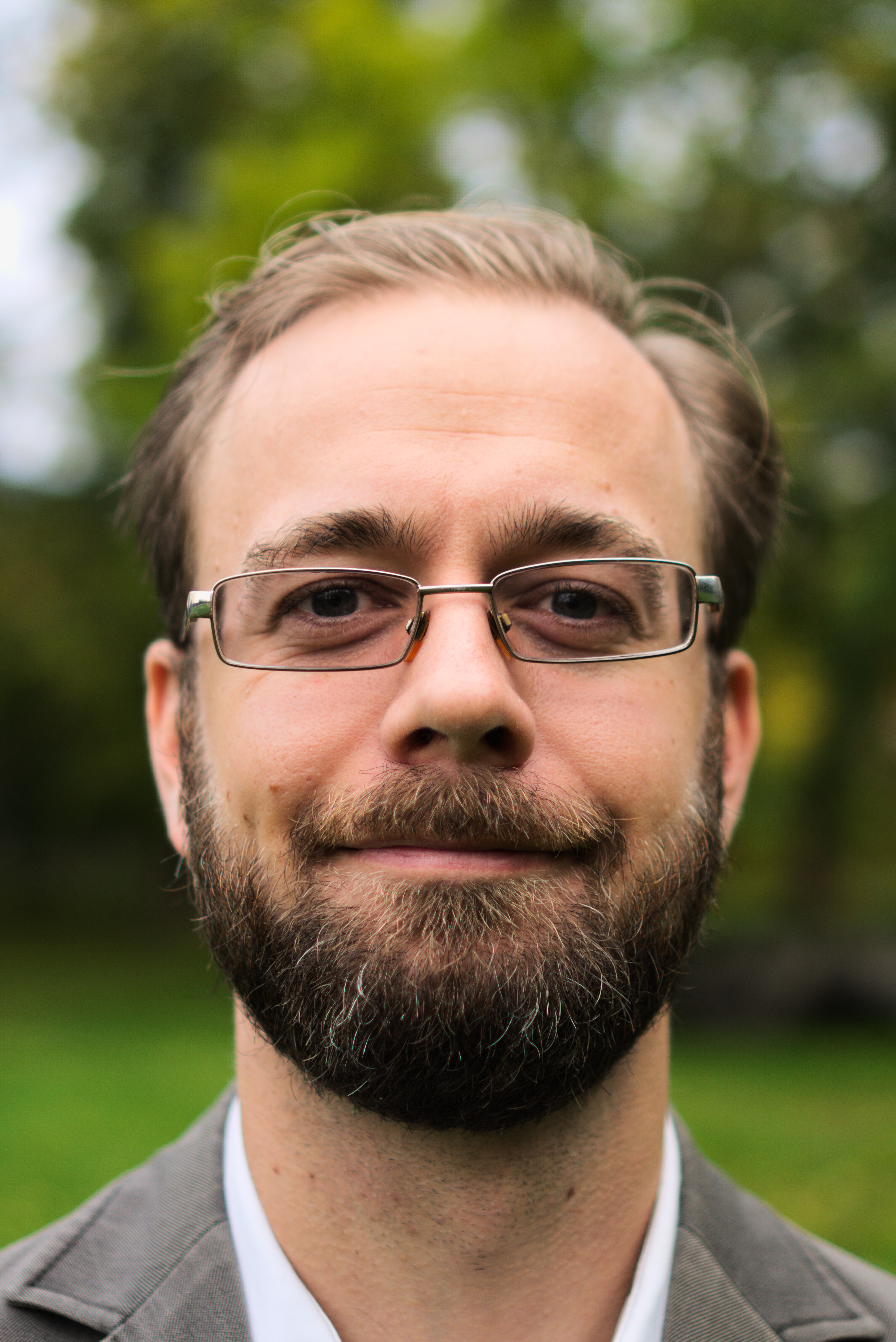}}]{Gustaf Hendeby} (Senior member, IEEE) received the M.Sc. degree in applied physics and electrical engineering in 2002 and the Ph.D. degree in automatic control in 2008, both from Linkoping University, Linkoping, Sweden. He is Associate Professor and Docent in the division of Automatic Control, Department of Electrical Engineering, Linkoping University. He worked as Senior Researcher at the German Research Center for Artificial Intelligence (DFKI) 2009–2011, and Senior Scientist at Swedish Defense Research Agency (FOI) and held an adjunct Associate Professor position at Linkoping University 2011–2015. His main research interests are stochastic signal processing and sensor fusion with applications to nonlinear problems, target tracking, and simultaneous localization and mapping (SLAM), and is the author of several published articles and conference papers in the area. He has experience of both theoretical analysis as well as implementation aspects. Dr. Hendeby was an Associate Editor for IEEE Transactions on Aerospace and Electronic Systems in the area of Target Tracking and Multisensor Systems 2018--2025, and is since 2025 a Senior Editor. In 2022 he served as general chair for the 25th IEEE International Conference on Information Fusion (FUSION) in Linkoping, Sweden.
\end{IEEEbiography}
\begin{IEEEbiography}
[{\includegraphics[width=1in,height=1.25in,clip,keepaspectratio]{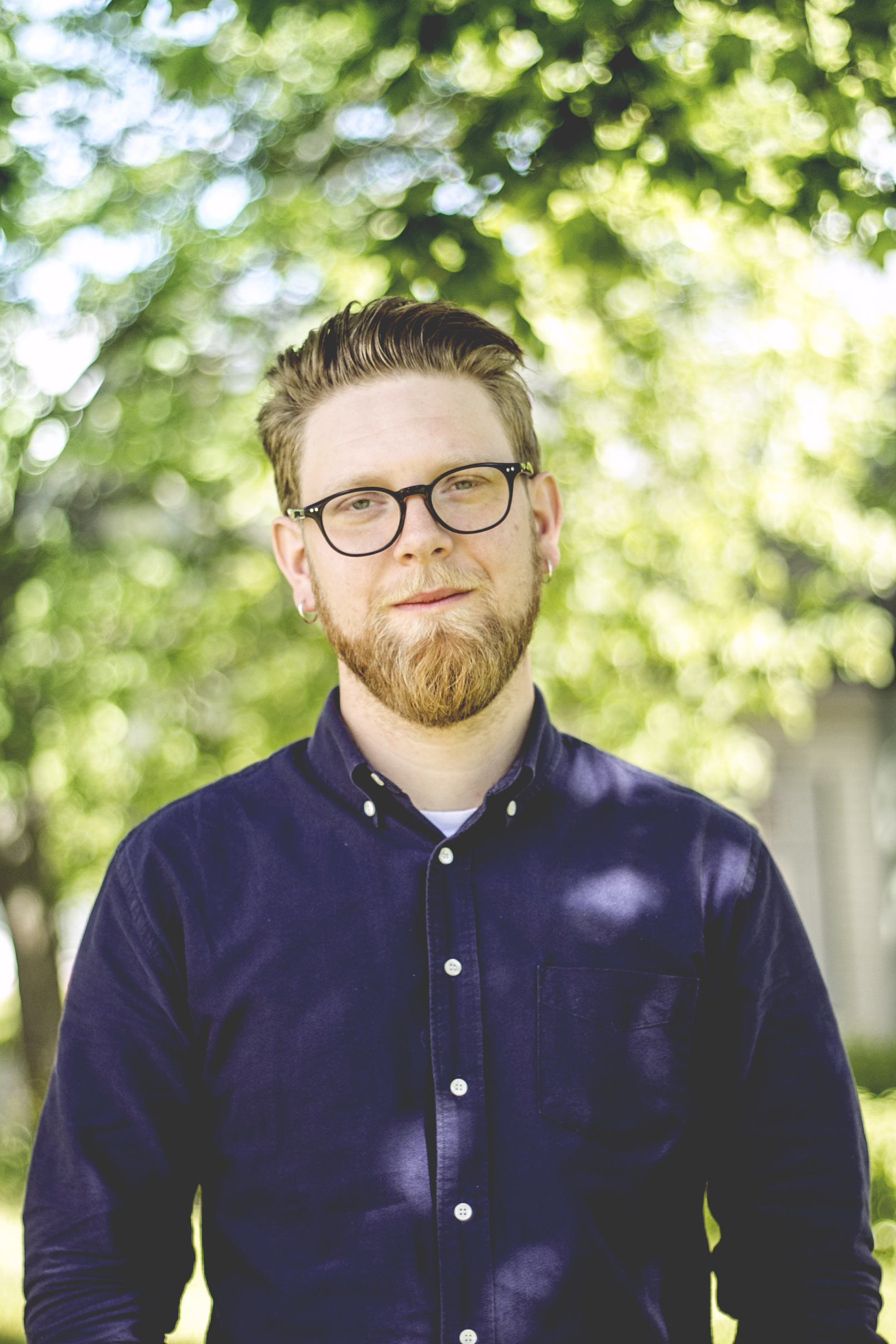}}]{Isaac Skog} (Senior Member, IEEE) received the B.Sc. and M.Sc. degrees in electrical engineering from the KTH Royal Institute of Technology, Stockholm, Sweden, in 2003 and 2005, respectively. In 2010, he received the Ph.D. degree in signal processing with a thesis on low-cost navigation systems. In 2009, he spent 5 months with the Mobile Multi-Sensor System Research Team, University of Calgary, Calgary, AB, Canada, as a Visiting Scholar and in 2011 he spent 4 months with the Indian Institute of Science, Bangalore, India, as a Visiting Scholar. Between 2010 and 2017, he was a Researcher with the KTH Royal Institute of Technology. He is currently an Associate Professor with Linköping University, Linköping, Sweden, and a Senior Researcher with Swedish Defence Research Agency (FOI), Stockholm, Sweden. He is the author and coauthor of more than 60 international journal and conference publications. He was the recipient of the Best Survey Paper Award by the IEEE Intelligent Transportation Systems Society in 2013.
\end{IEEEbiography}
\end{document}